\documentclass[Afour,sagev,times,doublespace]{sagej}

\usepackage{moreverb,url}
\usepackage{graphicx}

\usepackage{graphicx}

\usepackage{dblfloatfix} 
\usepackage{amsmath,mathtools}
\allowdisplaybreaks
\usepackage{adjustbox,booktabs,multirow}
\usepackage{booktabs,multirow,array}
\usepackage{multirow}
\usepackage[table]{xcolor}
\usepackage{colortbl}
\usepackage{graphicx} 
\usepackage{amssymb}       
\usepackage{caption}       
\usepackage{subcaption}    
\usepackage{tikz}          
\usepackage{float}         
\usepackage{geometry}
\usepackage{booktabs}  

\usepackage{amssymb}
\usepackage{needspace}
\usepackage{adjustbox}  
\usepackage{placeins}
\usepackage{booktabs} 
\usepackage{tabularx, booktabs}
\usepackage{graphicx} 
\usepackage{algpseudocode}
\usepackage[ruled,vlined]{algorithm2e}
\usepackage{fancyhdr}  
\SetAlgoVlined  
\SetAlgoNlRelativeSize{-1}  
\DontPrintSemicolon  
\usepackage{moreverb,url}

\usepackage[colorlinks=true,linkcolor=blue,citecolor=blue,urlcolor=blue]{hyperref}

\newcommand\BibTeX{{\rmfamily B\kern-.05em \textsc{i\kern-.025em b}\kern-.08em
T\kern-.1667em\lower.7ex\hbox{E}\kern-.125emX}}

\def\volumeyear{2025}

\begin{document}

\runninghead{Nabahirwa et al}

\title{An Empirical Study on the Robustness of YOLO Models for Underwater Object Detection}

\author{Edwine Nabahirwa\affilnum{1}, Wei Song\affilnum{1}, Minghua Zhang\affilnum{1}, Shufan Chen\affilnum{1}}

\affiliation{\affilnum{1} Shanghai Ocean University }







\begin{abstract}
Underwater object detection (UOD) remains a critical challenge in computer vision due to underwater distortions which degrade low-level features and compromise the reliability of even state-of-the-art detectors. While YOLO models have become the backbone of real-time object detection, little work has systematically examined their robustness under these uniquely challenging conditions. This raises a critical question: Are YOLO models genuinely robust when operating under the chaotic and unpredictable conditions of underwater environments?
In this study, we present one of the first comprehensive evaluations of recent YOLO variants (YOLOv8-YOLOv12) across six simulated underwater environments: low contrast, blur, noise, greenish color cast, bluish color cast, and clean-water enhancements. Using a unified dataset of 10,000 annotated images from DUO and Roboflow100, we not only benchmark model robustness but also analyze how distortions affect key low-level features such as texture, edges, and color. Our findings show that (1) YOLOv12 delivers the strongest overall performance but is highly vulnerable to noise, and (2) noise disrupts edge and texture features, explaining the poor detection performance in noisy images.
Class imbalance is a persistent challenge in UOD. Although classes with fewer samples appear harder to detect, attributes such as object size, shape, and color may make them easier to recognize. In contrast, dominant classes with complex visual features often achieve higher accuracy due to repeated training exposure. Experiments revealed that (3) image counts and instance frequency primarily drive detection performance, while object appearance exerts only a secondary influence.
Finally, we evaluated lightweight training-aware strategies: noise-aware sample injection, which improves robustness in both noisy and real-world conditions, and fine-tuning with advanced enhancement, which boosts accuracy in enhanced domains but slightly lowers performance in original data, demonstrating strong potential for domain adaptation, respectively. Together, these insights provide practical guidance for building resilient and cost-efficient UOD systems.
\end{abstract}


\maketitle

\section{Introduction}
Although object detection systems have made great progress on land, the underwater world still poses a mysterious and challenging environment for machines to understand. The medium of water introduces a host of visual degradations, such as light absorption, scattering, suspended particles, and spectral distortion, resulting in low contrast, motion blur, and severe color shifts \cite{er_research_2023}, \cite{xu_systematic_2023}, \cite{jian_underwater_2024}. These distortions degrade critical visual cues such as edges, texture, and shape, affecting the robustness and generalization capacity of even state-of-the-art object detectors \cite{jian_underwater_2024}, \cite{wang_refining_2025}.
The human eye may adapt to underwater settings through contextual reasoning, but machine learning models, especially those that rely on convolutional backbones, often fail to cope when presented with such environments. To successfully use deep learning models in real underwater tasks, such as tracking marine life, exploring the ocean, or inspecting underwater structures, we need to understand how these models handle the visual challenges associated with underwater environments \cite{jian_underwater_2024}.
YOLO (You Only Look Once) models have established themselves as the backbone of real-time object detection research and applications \cite{ali_yolo_2024}. From YOLOv8 to YOLOv12, enhancements in backbone networks (e.g., CSPDarknet, ELAN, PANet, CSPNeXt), detection heads, and attention modules have steadily pushed the frontier of detection accuracy and speed \cite{feng_ceh-yolo_2024}.

\begin{table*}[htbp]
\centering
\caption{Summary of Selected YOLO Variants: Architectural Innovations and Parameter Counts (in Millions)}
\begin{tabular}{p{1.8cm} p{2.5cm} p{3.3cm} p{5.4cm}}
\toprule
\textbf{Model} & \textbf{Parameters (M)} & \textbf{Key Components} & \textbf{Notable Innovations and Suitability} \\
\midrule
YOLOv8m \cite{yolov8_ultralytics} 
& 26.2 
& CSPDarknet, PANet, Anchor-free head 
& Baseline model with cross-stage partial connections and multi-scale fusion, efficient and suitable for real-time underwater detection. \\

YOLOv9c \cite{leonardis_yolov9_2025} 
& 25.5
& GELAN (CSPNet + ELAN), PANet, Anchor-free head 
& Enhances gradient flow and feature learning, improving expressiveness while retaining PANet structure. \\

YOLOv10m \cite{wang_yolov10_2024} 
& 15.4
& Transformer-enhanced CSP, Refined PANet, Improved head 
& Captures global context essential for detecting objects in variable underwater conditions, better robustness to size and orientation variations. \\

YOLO11m \cite{yolo11_ultralytics} 
&  20.1
& PANet + C3K2 + C2PSA, Lightweight separable head 
& Strengthens multi-scale fusion, improves small and occluded object detection with low computation, optimized for underwater clutter. \\

YOLOv12m \cite{tian_yolov12_2025} 
& 20.2
& R-ELAN, Area Attention (A2) 
& Leverages attention mechanisms for long-range context modeling, significantly boosting detection accuracy in complex scenes. \\
\bottomrule
\end{tabular}
\label{Table.1}
\end{table*}

YOLO models have been applied in various domains such as autonomous driving \cite{kang_improved_2024}, \cite{yue_wgs-yolo_2024}, \cite{li_yolo-vehicle-pro_2024}, surveillance \cite{bajpai_yolo_2024}, aerial imagery \cite{xiao_fbrt-yolo_2025}, medical imaging \cite{mohanty_automated_2025}, \cite{sobek_medyolo_2024}, \cite{bai_scc-yolo_2025}, agriculture \cite{patel_accurate_2024}, \cite{badgujar_agricultural_2024}, and industrial inspection \cite{raimundo_yolox-ray_2023}, which are ideal or moderately noisy conditions. However, a central question remains: \textbf{Are YOLO models genuinely robust when operating under the chaotic and unpredictable conditions of underwater environments?}

This study aims to fill an important gap by thoroughly evaluating how recent YOLO variants perform in challenging underwater environments, where factors like poor visibility, color distortion, and noise often disrupt object detection accuracy.

To explore how modern object detectors handle real underwater challenges, we combined two public datasets (DUO and Roboflow100), resulting in a unified set of 10,000 annotated images across four marine object classes. Since existing datasets rarely capture the full range of underwater distortions, we simulated six common visual conditions: low contrast, motion blur, greenish and bluish color shifts, noise, and clean-water enhancements \cite{purnima_devising_2025} (see Sections 2.2 and 2.3).
We then evaluated five recent YOLO models: YOLOv8m, YOLOv9c, YOLOv10m, YOLO11m, and YOLOv12m (see Table~\ref{Table.1}), by training them on original real-world data and testing them under each distortion type. This setup allowed us to systematically assess how well these models generalize and remain robust when faced with the unpredictable realities of underwater scenes.

Beyond just evaluating the overall performance of the model, we also took a closer look at how specific low-level image features are affected by noise, contrast enhancements for clean-water, and advanced enhancement. To do this, we analyzed texture using GLCM \cite{dash_glcm_2021}, \cite{armi_texture_2019}, edges using Sobel filters \cite{priyanka_implementation_2024}, and color using raw RGB values. This helps clarify which basic visual features are most impacted by underwater distortions, how these changes influence detection performance, and what strategies can be recommended to address them.

Class imbalance is a well-known challenge in underwater object detection. Although it may seem intuitive that classes with fewer examples are harder to detect \cite{chen_underwater_2024}, this is not always the case. Factors such as object size, shape, and color can make some classes easier to recognize, even if they are underrepresented. In contrast, a dominant class with complex visual features may still achieve high detection accuracy simply because the model has had more exposure to it during training. Taking these dynamics into account, this study investigates how the number of images containing a particular class and the overall frequency of its instances affect the detection performance, both for that class and in general.

To strengthen YOLOv12’s ability to handle underwater noise and distortion, we explore lightweight training-aware strategies. Specifically, we consider (i) small-sample injections, designed to expose the model to realistic variations without overwhelming it, and (ii) fine-tuning with advanced enhancement methods \cite{guo_underwater_2025}, aimed at adapting the model to feature-preserving image transformations. These approaches are expected to improve robustness and provide a practical pathway for domain adaptation, where a model trained in one underwater setting can be effectively transferred to another with limited additional data.

The contributions of this work are as follows:
\begin{enumerate}
    \item We conduct one of the first systematic robustness evaluations of recent YOLO models (YOLOV8, YOLOV9, YOLOV10, YOLO11, YOLOV12) under diverse and challenging underwater conditions, providing clear insights into why detection performance declines in noisy and distorted environments.  
    
    \item We establish a clear link between low-level feature degradation, covering texture, edges, and color, and detection accuracy through a targeted sensitivity analysis of critical underwater distortions such as noise. Furthermore, we show that class imbalance is primarily driven by image count and instance frequency, with object appearance playing only a secondary role.
    
   \item We demonstrate that lightweight training strategies can significantly improve robustness. Specifically, (i) small noise-aware data injection enhances performance under both noisy and original real-world underwater conditions, and (ii) fine-tuning with advanced enhancement improves performance in enhanced datasets, though with a slight reduction in original real-world accuracy. This highlights fine-tuning as an effective tool for domain adaptation, enabling pretrained YOLO models trained in one underwater environment to transfer effectively to another (e.g., lakes, oceans, shallow or deep waters) with limited additional data, offering a cost-efficient pathway toward practical deployment.
\end{enumerate}

\subsection{Research Questions and Scope}
Understanding the operational boundaries of object detectors in non-ideal environments is a critical step toward deploying AI systems in the real world. Although state-of-the-art YOLO models have demonstrated remarkable accuracy and speed in conventional benchmarks, their behavior under the complex and dynamic conditions of underwater environments remains largely underexplored. Unlike terrestrial scenes, underwater imagery is troubled by degradation factors such as light attenuation, color shifts, blur, and particulate noise, all of which disrupt low-level visual cues vital for object detection.

Although prior UOD research has focused heavily on refining model architectures tailored for underwater conditions such as backbone and attention layer modifications \cite{fayaz_underwater_2022}, \cite{nguyen_improve_2025} and preprocessing techniques \cite{chen_underwater_2024(1)}, relatively few studies delve into why model performance degrades under real-world underwater distortions.

This study addresses this gap by establishing a unified framework that evaluates detection robustness, investigates feature distortion sensitivity, analyzes class imbalance effects, and suggests possible strategies. To guide this investigation, we pose the following research questions:
\\
\textbf{RQ1}: What levels of robustness and accuracy do recent YOLO models demonstrate when exposed to challenging underwater conditions?
\\
\textbf{RQ2}: How are low-level visual features degraded in the most challenging situations?
\\
\textbf{RQ3}: How does reducing the number of training images and instance frequency of a dominant class affect both individual class detection performance and overall object detection accuracy in the presence of class imbalance in underwater datasets?
\\
These three questions form the backbone of our experimental design, detailed in Section 3. RQ1 benchmarks five modern YOLO models under six simulated distortion types, establishing a robustness profile. RQ2 goes beyond performance metrics by dissecting how specific distortions impact texture, edge, and color-based feature representations. Finally, RQ3 explores the dynamics of the dataset, how image counts, frequency, object shape, and visibility may influence detection accuracy in presence of class imbalance.

\section{Methodology}
\label{sec:methodology}

\subsection{Methodology Overview}
\begin{figure*}[htbp]
  \centering
  \includegraphics[width=\textwidth]{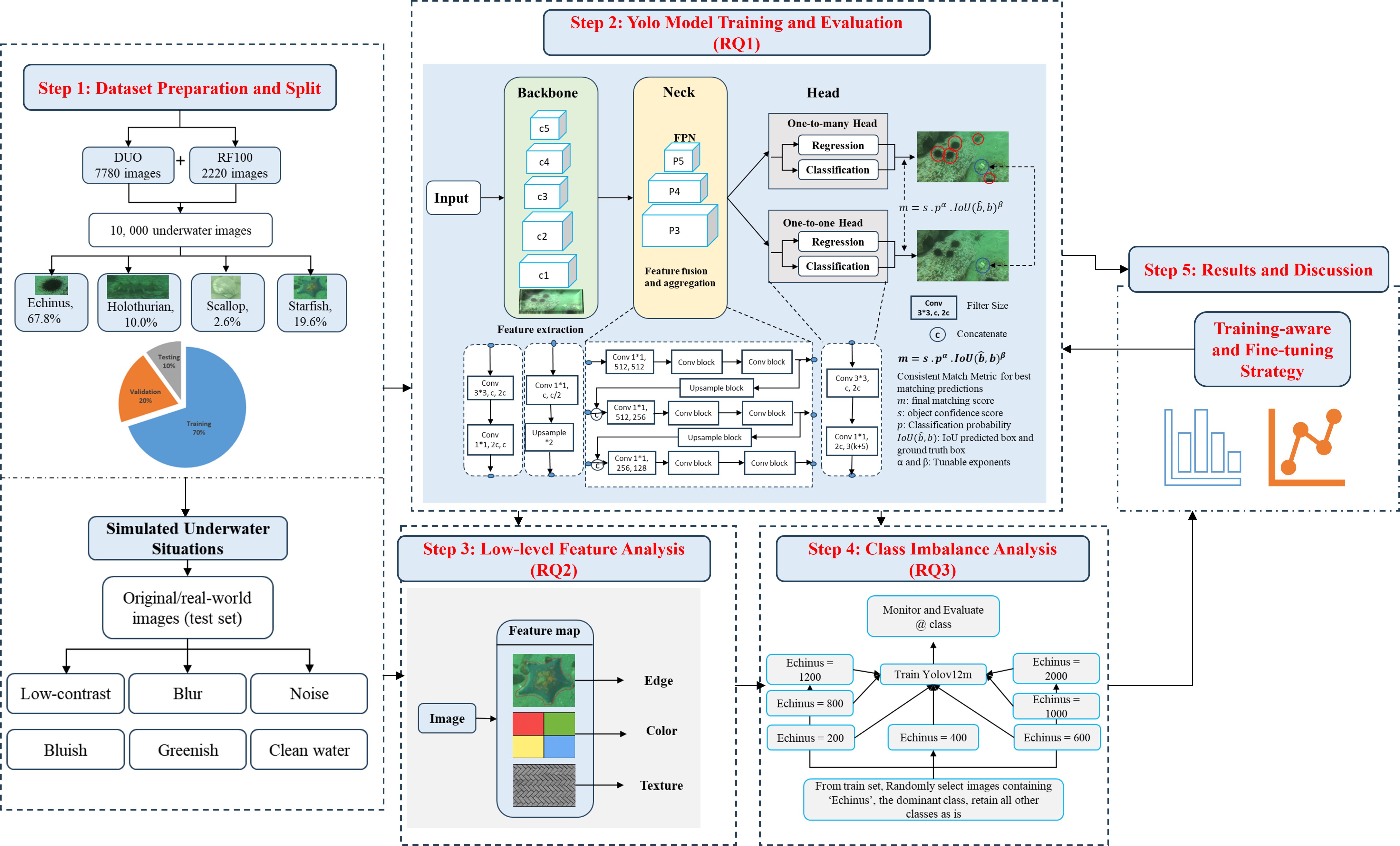}
  \caption{Overall Methodological Pipeline}
  \label{Fig.1}
\end{figure*}

Figure\ref{Fig.1} illustrates the methodological framework developed to examine the robustness and generalization of underwater object detection models under real-world and simulated distortions.
The process begins with Step 1: Dataset Preparation and Split, where the dataset is organized and various underwater environmental conditions are simulated, including low contrast, noise, blur, bluish, and greenish color casts, and clean-water, on the test set to replicate realistic underwater challenges.
In Step 2, Model Training and Evaluation addresses Research Question 1 (RQ1). We train five YOLO models (YOLOv8, YOLOv9, YOLOv10, YOLO11, and YOLOv12) on original, real-world underwater images and evaluate their performance across both real-world and distorted images. To balance performance and practicality, we utilize the medium variants (‘m’) of YOLOv8, YOLOv10, YOLO11, and YOLOv12. For YOLOv9, we select the compact variant (‘c’) as the medium variant was not available during our experiments; the ‘c’ variant provided comparable accuracy to the ‘m’ variants of the other models while offering lower computational cost.

Step 3: Low-Level Feature Analysis is conducted to answer RQ2. This step involves analyzing the effect of noise distortion, clean-water, and advanced enhancement effects on low-level visual features such as edge sharpness, color consistency, and texture clarity, using tools such as GLCM, Sobel filters and RGB statistics.
Step 4: Class Imbalance Analysis targets RQ3 to investigate how image counts, instance frequency, and maybe appearance of certain object classes influence detection performance.
Finally, Step 5: Results and Discussion brings everything together. Building on the insights from earlier stages, this step designs a small-sample injection strategy and finetuning strategy to improve the model’s robustness under distortion without compromising its accuracy on real-world underwater data.

\subsection{Dataset Construction and Split}

To build a diverse and representative dataset for underwater object detection, we merged 7,780 images from the DUO dataset \cite{liu_dataset_2021} and 2,220 images from the RF100 dataset \cite{ciaglia_roboflow_2022}, resulting in a combined set of 10,000 annotated underwater images. The dataset focuses on four common marine object classes: echinus, holothurian, scallop, and starfish. However, the distribution between classes is markedly unbalanced, where echinus accounts for roughly 67.8\% of all instances, followed by starfish (19.6\%), holothurian (10\%) and scallop (2.6\%) as shown in Table~\ref{Table.2}. This imbalance reflects common sampling biases found in real-world underwater environments and introduces challenges in achieving fair model generalization across classes. To support robust evaluation, the dataset was randomly split into 70\% training (7,000 images), 20\% validation (2,000 images) and 10\% testing (1,000 images).


\begin{table}[t]
\centering
\caption{Summary of Dataset Composition and Class Distribution}
\label{Table.2}
\footnotesize
\begin{adjustbox}{width=\columnwidth}
\begin{tabular}{@{}l l l r@{}}
\toprule
\textbf{Dataset} & \textbf{Sample Size} & \textbf{Classes} & \textbf{Instances} \\
\midrule
\multirow{4}{*}{DUO and RF100}
& \multirow{4}{*}{(7,780 + 2,220)}
& Echinus      & 56,816 \\
& & Holothurian & 10,019 \\
& & Scallop     & 3,745 \\
& & Starfish    & 17,808 \\
\bottomrule
\end{tabular}
\end{adjustbox}
\end{table}

\subsection{Simulated Underwater Environmental Conditions}
\begin{figure*}[htbp]
  \centering
  \includegraphics[width=\textwidth]{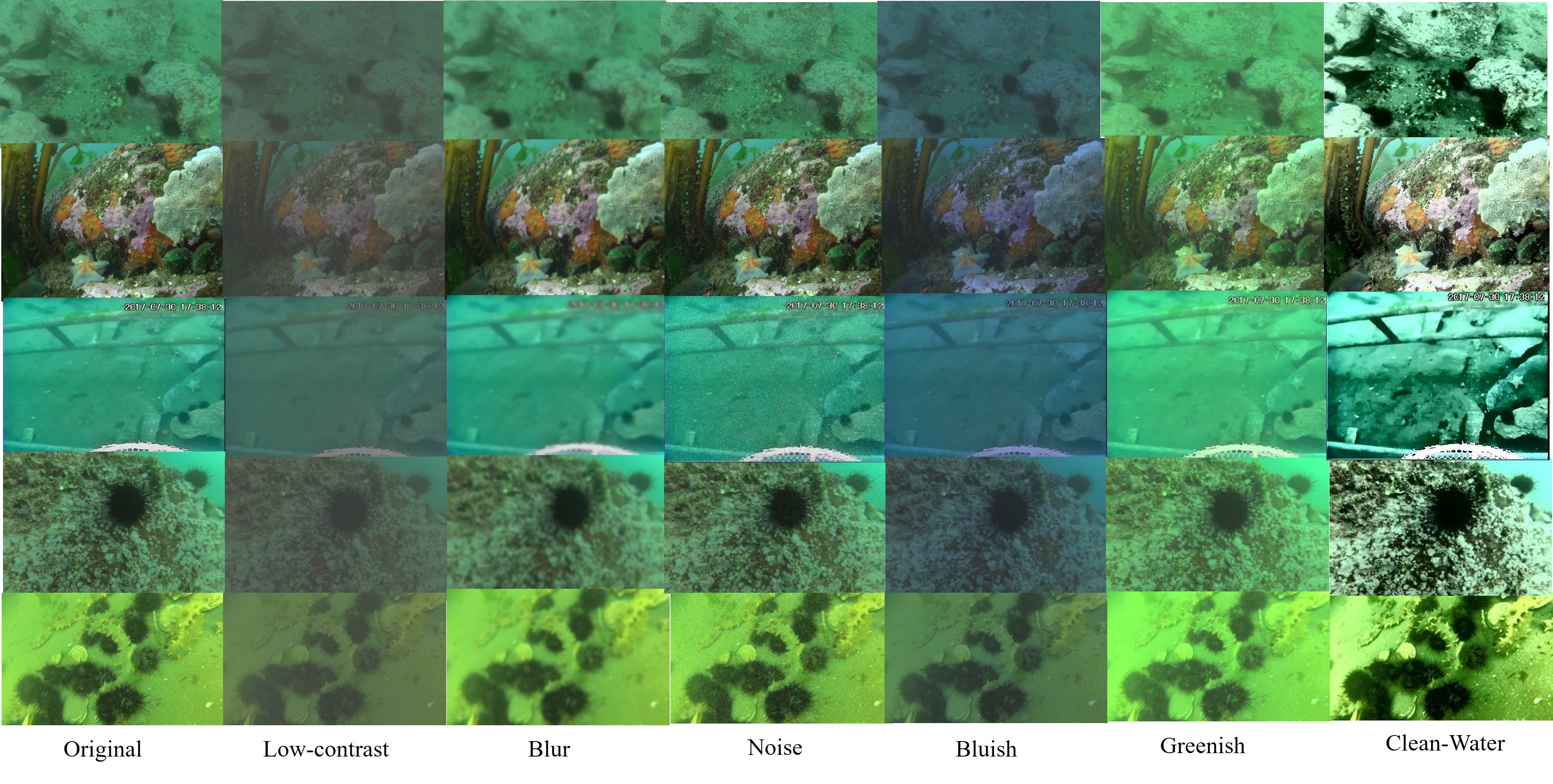}
  \caption{Visual comparison of original underwater images and six simulated distortion types (low contrast, blur, noise, bluish cast, greenish cast, and clean-water).}
  \label{Fig.2}
\end{figure*}

To evaluate the robustness of the model under real-world challenges, we simulate common underwater distortions, including low contrast, blur, noise, color shifts (bluish/greenish) and clean-water on the test set. These perturbations reflect typical degradation scenarios encountered in underwater imagery and are visually illustrated in Figure\ref{Fig.2}.

\subsubsection*{Low-Contrast Simulation}
To replicate the visual challenges posed by poor lighting and reduced contrast in underwater environments, we applied a two-step image transformation: global brightness attenuation followed by contrast compression. This procedure mimics the limited dynamic range and washed-out appearance caused by light absorption and scattering underwater. Let $I \in [0,255]^{H \times W \times 3}$ be the original RGB image, where $H$, $W$, and $3$ denote the height, width, and color channels. All pixel values are first converted to floating-point precision before transformation.

\textbf{1. Brightness Reduction:}
Global illumination is reduced using a brightness factor $\alpha \in (0,1)$:
\begin{align}
I_{\text{low-light}}(x,y,c)
  &= \min\!\bigl(\max\!\bigl(\alpha\, I(x,y,c),0\bigr),\,255\bigr),
  \label{eq:low_light}
\end{align}
where $\alpha=0.6$ in our implementation. This uniformly darkens the image, simulating dim underwater lighting.

\textbf{2. Contrast Compression:}  
Contrast is reduced using a factor $\beta \in (0,1)$:
\begin{equation}
\begin{split}
I_{\text{low-contrast}}(x,y,c) 
   &= \min\!\Big( \max\!\big( (I_{\text{low-light}}(x,y,c)\\
   -128)\cdot \beta + 128,\,0 \big), 
   &\quad 255 \Big),
\end{split}
\label{eq:low_contrast}
\end{equation}
with $\beta = 0.6$.
This transformation pulls pixel intensities toward the mid-tone (128), flattening the histogram and simulating the low-contrast, hazy appearance typical of underwater imagery.
The resulting images serve as controlled test inputs to assess model robustness under low-contrast conditions.

\subsubsection*{Blurred Image Simulation}

Underwater environments often introduce blur \cite{li_underwater_2024} due to movement of water, light scattering, or limitations in underwater imaging systems. Blurring distorts object boundaries and fine textures, posing challenges for object detection algorithms. To simulate this degradation, Gaussian blur was applied to the test set.

Gaussian blur is a smoothing technique that reduces image sharpness by convolving the input image with a Gaussian kernel. Let $I$ be the original image and $G$ the Gaussian kernel of size $k \times k$, the blurred image $I_{\text{gauss}}$ is obtained as:
\begin{equation}
I_{\text{gauss}}(x, y) = \sum_{i=-k}^{k} \sum_{j=-k}^{k} G(i, j) \cdot I(x - i, y - j)
\end{equation}
In our implementation, we used a $19 \times 19$ kernel to ensure a strong blurring effect representative of the distortion typically observed in turbid or fast-moving underwater scenes.

\subsubsection*{Noise Simulation}

Underwater images are frequently affected by various types of noise due to low light conditions, sensor limitations, and suspended particles in water \cite{lu_underwater_2023}. To simulate this real-world degradation, we applied a combination of Gaussian noise and salt-and-pepper noise to the test set.

\textbf{1. Gaussian Noise:}

Gaussian noise introduces random variations in pixel intensity, following a normal distribution. Given an image $I$ and Gaussian noise $N \sim \mathcal{N}(\mu, \sigma^2)$, the noisy image $I_{\text{gauss}}$ is computed as:
\begin{equation}
I_{\text{gauss}}(x, y, c) = \text{clip}\left(I(x, y, c) + N(x, y, c),\ 0,\ 255\right)
\end{equation}
where $\mu = 0$ and $\sigma = 10$ in our experiment. This simulates electronic noise and minor disturbances typical of underwater camera sensors.

\textbf{2. Salt-and-Pepper Noise:}

Salt-and-pepper noise randomly replaces some pixel values with the extremes (0 for black, 255 for white), mimicking artifacts caused by transmission errors or image corruption. The probability of salt ($p_s$) and pepper ($p_p$) noise is set to 0.005. For an image with $N$ total pixels, the number of salt and pepper pixels are:
\begin{equation}
n_s = N \cdot p_s,\quad n_p = N \cdot p_p
\end{equation}
These noisy pixels are randomly assigned as:
\begin{equation}
I_{\text{salt}}(x_i, y_i) = 255,\quad I_{\text{pepper}}(x_j, y_j) = 0
\end{equation}

\textbf{3. Combined Noise:}

The final noisy image $I_{\text{noisy}}$ is obtained by sequentially applying Gaussian noise followed by salt-and-pepper noise:
\begin{equation}
I_{\text{noisy}} = \text{SaltAndPepper}\left(\text{Gaussian}(I)\right)
\end{equation}
This combined noise simulation captures the intensity fluctuations and impulse noise commonly found in degraded underwater images.

\textbf{4. Gaussian-diffusion Noise Simulation:}

In this type of noise we simulate a gradual corruption of the image by incrementally adding small amounts of Gaussian noise over multiple steps. This technique mimics the progressive degradation of visual information and serves as a simple forward step inspired by diffusion-based generative models \cite{qi_not_2024}. In this experiment, noise was applied to a subset of test set images to evaluate model robustness against iteratively intensified distortions.
Let $I \in [0, 1]^{H \times W \times C}$ be the normalized input image tensor and let $T$ be the total number of noise steps. At each step $t$, Gaussian noise is added to the image as follows:
\begin{equation}
I^{(t)} = \text{clip}\left(I^{(t-1)} + \epsilon^{(t)},\ 0,\ 1\right),\quad \epsilon^{(t)} \sim \mathcal{N}(0,\ \sigma^2)
\end{equation}
where:
- $I^{(0)} = I$ (the original image),
- $\sigma = 0.05$ is the noise scale,
- $\epsilon^{(t)}$ is Gaussian noise sampled at each iteration,
- $\text{clip}(\cdot)$ constrains pixel values to the range $[0, 1]$.
After $T = 3$ iterations, the final noisy image $I^{(T)}$ represents a softly degraded version of the original image, with subtle texture and intensity corruption across all channels.

\subsubsection*{Bluish Environment Simulation}

In natural underwater settings, the absorption of light at different wavelengths leads to a dominant bluish or cyan color cast, especially at greater depths. Red light is first absorbed, while blue light penetrates deeper, resulting in significant color distortion in underwater images \cite{alenezi_underwater_2022}. 
Underwater color cast arises from wavelength-dependent attenuation (red $\!>\!$ green $\!>\!$ blue) and additive veiling light from the water body. We simulate this with the classical formation model:

\begin{equation}
I_c(x,y) = J_c(x,y)\,t_c + A_c\,(1-t_c), \qquad 
t_c = \exp(-\beta_c\,d),
\label{eq:uw_model}
\end{equation}
where $J$ is the input (scene radiance) image, $I$ the rendered underwater image, $c\!\in\!\{R,G,B\}$, $d$ is an effective water depth (or haze thickness), $\beta_c$ are per-channel attenuation coefficients with $\beta_R>\beta_G>\beta_B$, and $A=(A_R,A_G,A_B)$ is the ambient waterlight (bluish) color.
For images in $[0,255]$, we operate in linear RGB, then convert back to sRGB and clip:

\begin{equation}
\begin{split}
\tilde{J} &= \mathrm{lin}\!\left(\tfrac{J}{255}\right), \\
\tilde{I}_c(x,y) &= \tilde{J}_c(x,y)\,e^{-\beta_c d}
    + A_c\!\left(1-e^{-\beta_c d}\right), \\
I &= 255\,\mathrm{srgb}\!\bigl(\mathrm{clip}(\tilde{I},0,1)\bigr),
\end{split}
\label{eq:uw_srgb}
\end{equation}
A practical parameter set that yields a realistic bluish/cyan cast is, e.g.,

\begin{equation}
\begin{split}
\beta_R &= 0.12, \quad \beta_G = 0.07, \quad \beta_B = 0.03, \\
A &= (0.05,\,0.10,\,0.25),
\end{split}
\label{eq:uw_params}
\end{equation}
with $d$ in $[0.5,2.0]$ controlling cast strength. Larger $d$ or larger $\beta_R$ deepen the blue tint naturally, while $A_B>A_G>A_R$ enforces bluish veiling.

\subsubsection*{Greenish Environment Simulation}

In shallow coastal waters, underwater imagery often exhibits a greenish hue due to climate change and the presence of phytoplankton, algae, and suspended organic matter \cite{cael_global_2023}. These elements alter the spectral properties of light, resulting in a dominant green tint and affecting both color fidelity and object visibility. To replicate this visual distortion, we follow the same wavelength–dependent attenuation and veiling-light formulation described in the bluish environment simulation (Equations~\ref{eq:uw_model}--\ref{eq:uw_params}), but adapt the attenuation coefficients and ambient color to emphasize green rather than blue. Specifically, we use

\begin{equation}
\begin{split}
\beta_R &= 0.12, \quad \beta_G = 0.04, \quad \beta_B = 0.08, \\
A &= (0.05,\,0.20,\,0.10),
\end{split}
\label{eq:uw_green}
\end{equation}

so that red is still strongly absorbed, blue moderately suppressed, and the veiling light shifts toward green. This parameterization yields a realistic greenish underwater cast representative of algae- or plankton-rich waters.

\subsubsection*{Clean-water Simulation}

In this study, clean water refers to underwater scenes with high clarity and minimal distortions such as shallow tropical seas. To simulate this, we applied a simple two-step process: contrast adjustment followed by soft sharpening. This was done to make the object boundaries more noticeable and to improve the ease with which features can be detected. We chose this method to test whether even simple improvements in visual quality could help the models generalize better in different underwater conditions.
Let $I \in [0, 255]^{H \times W \times 3}$ be the original image. First, contrast is enhanced using a linear transformation centered around the mid-tone (128):
\begin{equation}
\begin{split}
I_{\text{contrast}}(x, y, c) 
  &= \text{clip}\Big( (I(x, y, c) - 128)\cdot \gamma + 128, \\
  &\quad 0,\,255 \Big)
\end{split}
\label{eq:contrast}
\end{equation}
where $\gamma = 1.2$ is the contrast amplification factor.
Next, sharpening is applied using a 2D convolution with a soft sharpening kernel $K$:
\begin{equation}
K = 
\begin{bmatrix}
0 & -0.25 & 0 \\
-0.25 & 2 & -0.25 \\
0 & -0.25 & 0
\end{bmatrix}
\end{equation}
The final clear-water image is computed as:
\begin{equation}
I_{\text{enhanced}}(x, y) = I_{\text{contrast}}(x, y) * K
\end{equation}
where $*$ denotes convolution. This kernel boosts the center pixel while slightly suppressing its neighbors, thereby sharpening edges without introducing harsh artifacts.

\subsection{Model Training and Evaluation \textbf{RQ1}}

This experiment aims to assess and compare the performance of five advanced YOLO models: YOLOv8m, YOLOv9c, YOLOv10m, YOLO11m and YOLOv12m under both real-world and simulated challenging underwater conditions. The main objective is to understand how each model handles distortions commonly found in underwater imagery and to evaluate their ability to generalize beyond real-world training data.
To ensure a fair comparison, all models are first trained on the same original real-world dataset. Once trained, they are tested on a test set available in both original and artificially distorted forms.
This setup allows us to explore how different YOLO versions perform when exposed to degraded environments. We evaluated them using standard metrics, including mean Average Precision (mAP), precision, recall, and per-class detection robustness, to identify which models are most resilient to underwater challenges.

\subsection{Feature-Level Sensitivity Analysis \textbf{RQ2}}

In our analysis of model robustness under RQ1, we noticed significant performance drops, particularly under noisy conditions that could not be fully explained by model architecture alone. This led us to conduct a deeper investigation at the feature level, with the aim of understanding how specific distortions affect the visual cues that detection models rely on.
In addition to evaluating noise, we included clean-water and enhanced images to better understand the influence of visual restoration. Saleem et al. \cite{saleem_understanding_2025} emphasize that the type of enhancement technique and its interaction with the detection model play a critical role in determining object detection performance. In this context, recent advances in deep learning-based enhancement methods have demonstrated promising results, with certain approaches significantly boosting detection accuracy \cite{awad_beneath_2024}. Building on this, our analysis focuses on images enhanced using HybSense \cite{guo_underwater_2025}, a new deep learning technique known for its effective visual restoration in underwater imagery.

HybSense is a deep learning-based underwater image enhancement method that employs a U-shaped encoder-decoder architecture called GuidedHySensUIR, designed to restore color fidelity and detail in degraded underwater scenes. At its core, the model processes the input RGB image along with a color balance prior, which helps guide global color correction. The architecture includes three key components: (1) Feature Extractor which extracts shallow visual features from the original image. (2) Feature Contextualizer that is positioned at the bottleneck, responsible for leveraging color balance priors to capture long-range dependencies and enhance color representation using transformer-style attention mechanisms. (3) Detail Restorer and Scale Harmonizer modules, located in the decoder path. The Detail Restorers enhance image textures at multiple scales, while Scale Harmonizers fuse multi-resolution features using channel-wise concatenation and upsampling/downsampling operations.

To assess the impact of these distortions on core visual features, we selected a random subset of 100 real-world underwater images from the training set and created four versions for each: (1) the original image, (2) a noisy version, (3) a clean-water version, and (4) an enhanced version. For each version, we analyzed the following low-level features: (i) texture using the Gray Level Co-occurrence Matrix (GLCM), (ii) edges using the Sobel operator, and (iii) color using raw RGB distributions.
This analysis helped reveal how distortions affect the structural integrity of underwater scenes and allowed us to better understand their impact on detection performance. Each type of characteristic was computed across the image sets, and the average response \(\bar{x}\) of each characteristic was compared between:
\[
\text{Original vs. Noisy} \quad \text{and} \quad \text{Original vs. Enhanced}
\]
The degradation of features \(\Delta_f\) for each set was quantified as:
\begin{equation}
\Delta_f = \left| \bar{x}_{\text{original}} - \bar{x}_{\text{distorted}} \right|
\end{equation}
This comparison helped identify which visual cues are most affected by underwater noise and clean-water and enhancement, and which features may contribute to the drop or improvement in detection accuracy. The conclusions drawn from this analysis informed our understanding of the feature types the models rely on and how underwater distortions weaken or improve model perception.

\subsubsection*{Texture Feature Extraction using GLCM}

Texture features provide crucial information for distinguishing object surfaces from the background in underwater imagery, especially under challenging distortions such as noise or color degradation. In this study, we used the GLCM \cite{dash_glcm_2021} to quantify the textural properties, due to its ability to capture spatial relationships between pixel intensities and its widespread use in image analysis \cite{zubair_grey_2024}.
Given a grayscale image $I \in \mathbb{R}^{H \times W}$, the GLCM $P(i, j)$ is calculated for a pair of pixels at a fixed spatial offset and angle (in our case, 1 pixel at $0^\circ$). From the normalized GLCM, we extracted four statistical features.

    \textbf{Contrast} Measures local intensity variation:
    \begin{equation}
    \text{Contrast} = \sum_{i,j} (i - j)^2 \cdot P(i, j)
    \end{equation}

    \textbf{Correlation} Measures the linear dependency between gray levels:
    \begin{equation}
    \text{Correlation} = \sum_{i,j} \frac{(i - \mu_i)(j - \mu_j) \cdot P(i, j)}{\sigma_i \cdot \sigma_j}
    \end{equation}

    \textbf{Energy} Measures textural uniformity (also known as Angular Second Moment):
    \begin{equation}
    \text{Energy} = \sum_{i,j} P(i, j)^2
    \end{equation}

    \textbf{Homogeneity} Measures the closeness of element distribution to the GLCM diagonal:
    \begin{equation}
    \text{Homogeneity} = \sum_{i,j} \frac{P(i, j)}{1 + |i - j|}
    \end{equation}

To analyze how distortions affect texture, we computed these features for each image in three regions: the full image, the object region (defined using bounding boxes), and the background region (complement of the object mask). The features were extracted for each version of the image: original, clean-water, noisy, and enhanced. The results were then compared to assess the impact of distortion on texture cues critical for object detection.

\subsubsection*{Edge Feature Extraction using Sobel Operator}

Edge features are vital for object detection models, particularly in underwater scenes, where visibility is reduced and object boundaries are often blurred. To quantify the presence and sharpness of edges, we employed the Sobel operator to compute edge density for objects across original, noisy, and enhanced images.

The Sobel operator \cite{priyanka_implementation_2024} detects edges by approximating the gradient of image intensity. For a given grayscale image $I(x, y)$, horizontal and vertical gradients are computed as:
\begin{equation}
G_x = \frac{\partial I}{\partial x}, \quad G_y = \frac{\partial I}{\partial y}
\end{equation}
These gradients are estimated via convolution with Sobel kernels:
\begin{equation}
G_x = I * S_x, \quad G_y = I * S_y
\end{equation}
The edge magnitude at each pixel is then given by:
\begin{equation}
G(x, y) = \sqrt{G_x(x, y)^2 + G_y(x, y)^2}
\end{equation}
To quantify edge sharpness, we define the \textbf{edge density} as the proportion of pixels within a detected object whose edge magnitude exceeds a given threshold $\tau$ (empirically set to 100):
\begin{equation}
\text{Edge Density} = \frac{1}{N} \sum_{x,y} \mathbb{1}\left(G(x, y) > \tau \right)
\end{equation}
where $N$ is the total number of pixels in the object region and $\mathbb{1}(\cdot)$ is the indicator function.
This metric was computed for every labeled object instance in three versions of each image: original, noisy, and enhanced. Sobel was chosen because of its simplicity, computational efficiency, and sensitivity to intensity transitions, making it effective for measuring how distortions affect object boundaries in underwater imagery.

\subsubsection*{Color Feature Extraction Using Raw RGB Statistics}
To assess the impact of underwater distortions on color-related features, we computed raw color statistics, including mean RGB values and 3D RGB histograms, to capture traditional color distribution properties from object regions across original, noisy, clean-water, and enhanced image sets. 
\text{Mean RGB Value of Object Crop}
Let the cropped image region be \( I_{\text{crop}} \in \mathbb{R}^{H \times W \times 3} \). The mean RGB vector is computed as:

\begin{equation}
\mu_{\text{RGB}} = \left[ \frac{1}{N} \sum_{n=1}^{N} R_n,\ \frac{1}{N} \sum_{n=1}^{N} G_n,\ \frac{1}{N} \sum_{n=1}^{N} B_n \right]
\end{equation}
where \( N = H \times W \), and \( R_n, G_n, B_n \) are the red, green, and blue channel values for each pixel \( n \) in the object crop.

\subsection{ Class Imbalance and influential class factors that affect detection \textbf{RQ3} }
\label{subsec:Class Imbalance analysis}

\begin{figure*}[htbp]
    \centering
    \includegraphics[width=\textwidth]{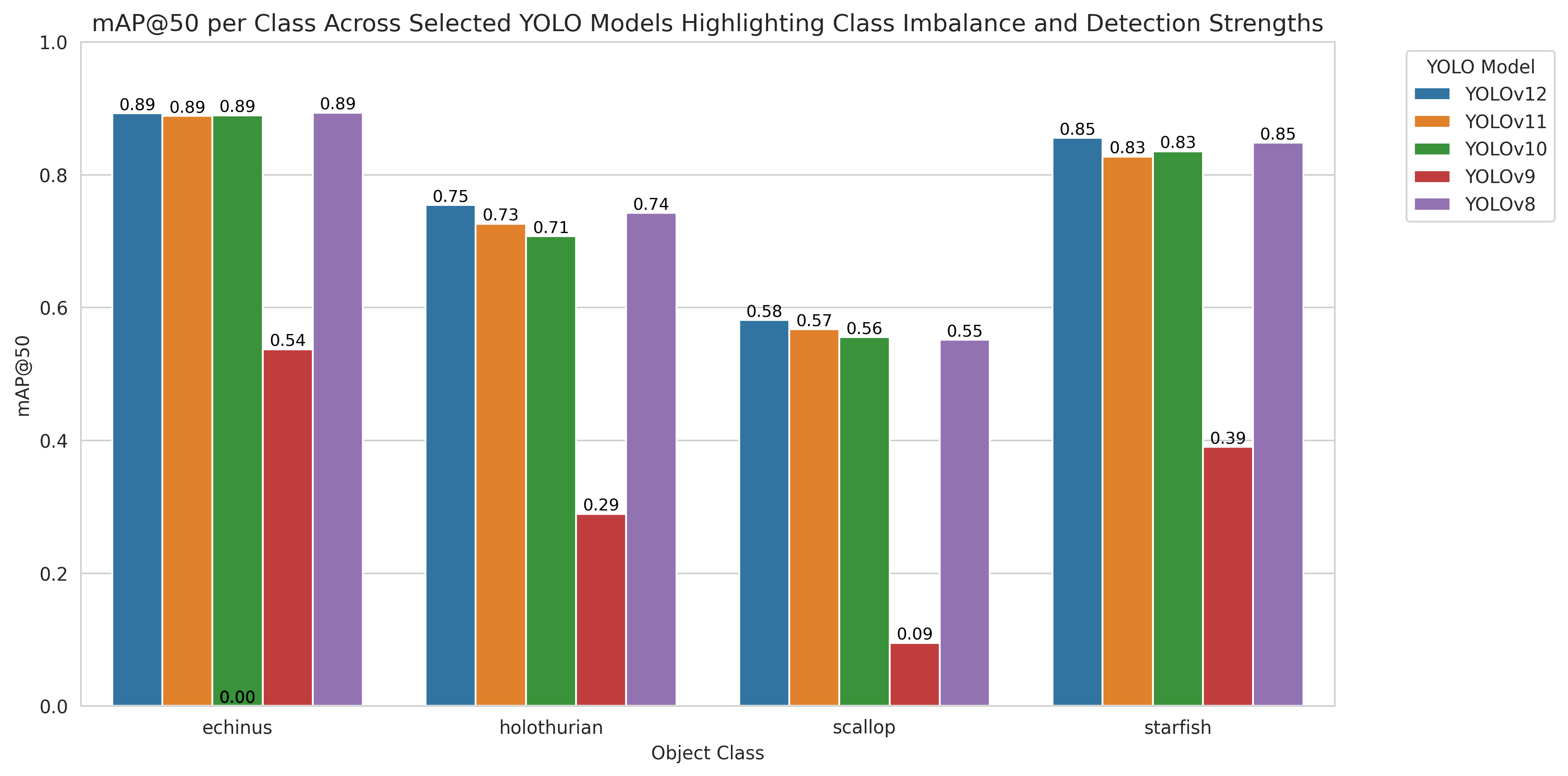}
    \caption{Comparison of mAP@50 across four object classes (echinus, holothurian, scallop, starfish) for YOLOv12, YOLO11, YOLOv10, YOLOv9, and YOLOv8 models on test (unseen) samples of the base real-world underwater dataset.}
    \label{fig.3}
\end{figure*}

During the initial stages of our experiments, we observed an interesting trend: although Echinus was the most dominant class in the dataset, accounting for approximately 67.8\% of all annotated instances, it consistently led in detection performance, as expected. However, what stood out was that Starfish, despite representing only 19.8\% of the data, occasionally achieved equally competitive mean Average Precision (mAP) scores (see Figure \ref{fig.3}).

These observations prompted further reflection on the relationship between data frequency and model performance, raising an important question: \textbf{Is having more examples in the training set enough to guarantee better detection performance?}

To explore this further, we conducted a controlled experiment to observe how gradually increasing the number of training images containing Echinus would influence its detection performance. We created seven subsets of the training data, each containing a different number of Echinus-containing images: 200, 400, 600, 800, 1000, 1200, and 2000, referred to as Subsets 1 through 7, respectively. The core idea was to vary only the number of Echinus instances, while keeping other conditions as consistent as possible.
However, since many images include multiple object types, the number of other classes in these subsets may have changed slightly due to the randomized selection process. While we did not intentionally adjust the distribution of the other classes, some minor fluctuations were inevitable. These slight variations were minimal and did not substantially affect the overall class distribution in the dataset (see Table~\ref{Table.3} for details). To maintain fairness in comparison, a separate YOLOv12m model was trained on each subset.

The goal is to track how detection metrics like precision, recall, and mAP change not just for Echinus, but also for the other classes, as we gradually increased the number of Echinus images. By keeping the variety of object instances constant and only adjusting how often Echinus appeared, we aimed to isolate the impact of image quantity from other factors like object shape or visibility.
Through this experiment, we sought to better understand whether detection challenges in underwater settings are mainly due to having fewer examples of some objects, or if they are also influenced by how clearly an object appears in an image. The insights gained here help inform smarter strategies for balancing datasets and improving model training.

\begin{table*}[htbp]
\centering
\caption{Image distribution (in \%) and instance counts per training subset. Seven subsets were created by varying the number of Echinus-containing images (200, 400, 600, 800, 1000, 1200, and 2000), referred to as Subsets 1–7, respectively.}
\label{Table.3}
\footnotesize
\setlength{\tabcolsep}{6pt} 
\begin{tabular}{clccccc r}
\toprule
\textbf{Subset} & \textbf{Counts} & \textbf{Echinus} & \textbf{Holothurian} & \textbf{Scallop} & \textbf{Starfish} & \textbf{Total} \\
\midrule
1 & Images (\%)  & 14.2  & 57.0 & 20.1 & 57.3 & 1406 \\
  & Instances    & 1168  & 1626 & 957  & 2390 & 6141 \\
2 & Images (\%)  & 24.9  & 56.2 & 18.9 & 58.2 & 1606 \\
  & Instances    & 2494  & 1836 & 990  & 2742 & 8062 \\
3 & Images (\%)  & 33.2  & 55.1 & 17.2 & 58.0 & 1806 \\
  & Instances    & 3823  & 2056 & 1015 & 3043 & 9937 \\
4 & Images (\%)  & 39.8  & 52.6 & 16.3 & 58.0 & 2006 \\
  & Instances    & 5639  & 2177 & 1052 & 3594 & 12462 \\
5 & Images (\%)  & 45.1  & 52.7 & 15.2 & 57.3 & 2206 \\
  & Instances    & 6754  & 2400 & 1101 & 3792 & 14047 \\
6 & Images (\%)  & 51.5  & 51.8 & 14.3 & 57.7 & 2406 \\
  & Instances    & 7970  & 2566 & 1112 & 4151 & 15799 \\
7 & Images (\%)  & 49.4  & 51.3 & 12.9 & 59.3 & 3206 \\
  & Instances    & 13593 & 3527 & 1394 & 5674 & 24188 \\
\bottomrule
\end{tabular}
\end{table*}

\subsection{Evaluation Metrics}
To fairly evaluate model performance across all experiments, we used standard object detection metrics such as mean average precision (mAP@50 and mAP@50:95), precision, and recall. These metrics were computed for each object class and under every distortion condition, giving us a well-rounded view of how accurate, reliable, and confident the models were in different underwater scenarios.
For analyzing image quality, especially when assessing enhanced images, we relied on widely used metrics such as PSNR (Peak Signal-to-Noise Ratio), SSIM (Structural Similarity Index), UIQM (Underwater Image Quality Measure), UCIQE (Underwater Color Image Quality Evaluation), and URanker.

\section{Results and Discussions}

\subsection{Implementation Details}

All model training and evaluation were performed using the PyTorch framework within the Google Colab environment, utilizing NVIDIA L4 GPUs. To ensure a fair and consistent comparison across all experiments, training settings were standardized. Each model was trained for a maximum of 60 epochs with an input resolution of 800×800 pixels. These uniform configurations facilitated reliable and reproducible comparisons between different models and experimental setups.

\subsection{Overall Model Robustness Across Distortions (RQ1)}

\paragraph{Experimental Recap} 
This evaluation investigates how five selected recent YOLO models perform under a variety of simulated underwater visual distortions throughout the test set, which comprises completely unseen data to measure true generalization ability.

\paragraph{Result Observations and interpretation:}

\begin{figure*}[htbp]
    \centering
    \includegraphics[width=\textwidth]{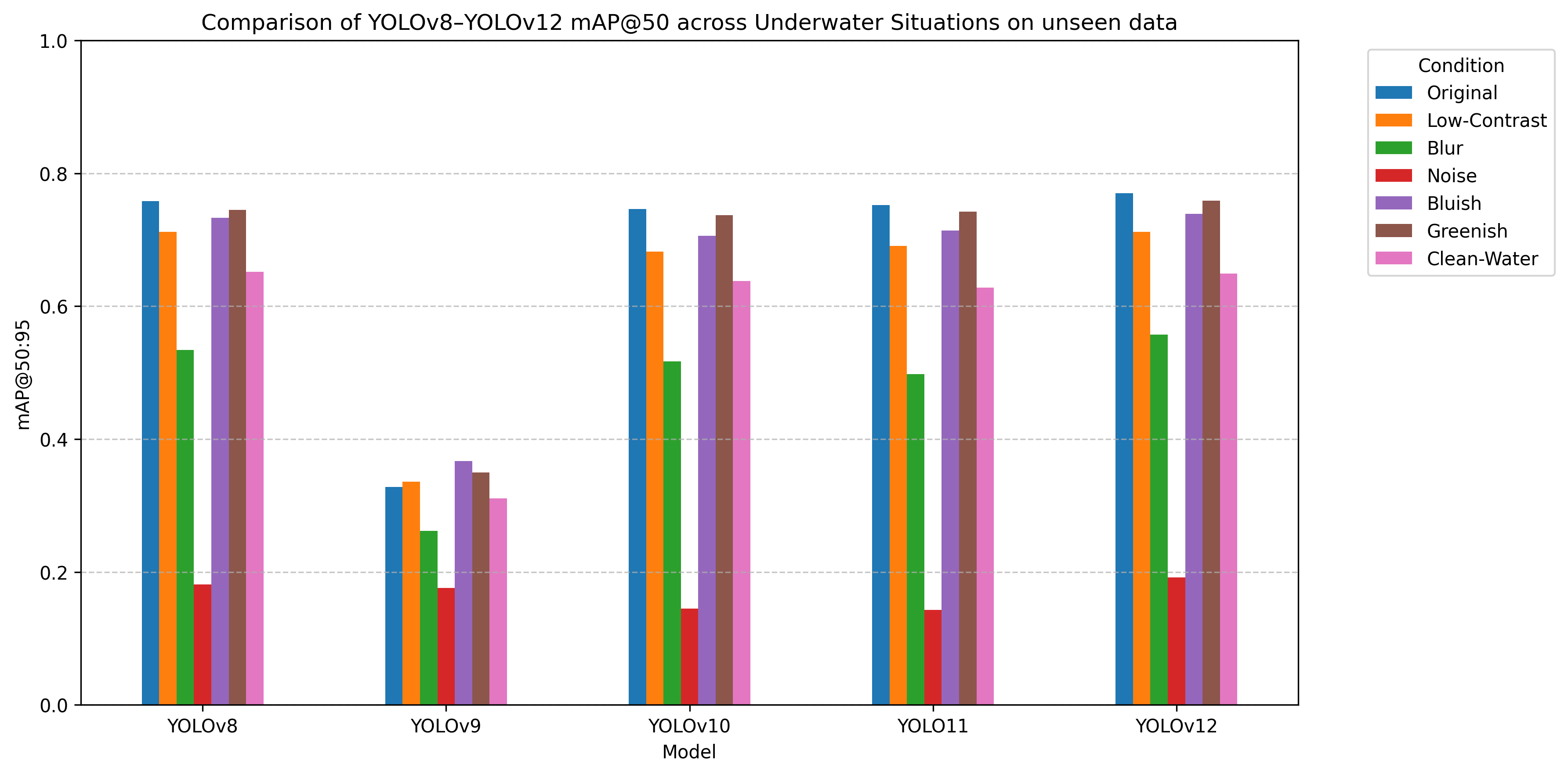}
    \caption{Comparison of YOLOv8 to YOLOv12 performance across seven underwater image conditions (Original, Low-Contrast, Blur, Noise, Bluish, Greenish, and Clean-water) using the mAP@50:95 metric on completely unseen data (test set).}
    \label{fig.4}
\end{figure*}

As seen in Table~\ref{Table.4} and  Figure\ref{fig.4}, all models were trained solely on real-world images, yet they exhibit relatively strong performance across most types of distortion. The test results reveal how the models truly generalize to new environments and image distributions.
In the original conditions, YOLOv12m scored the highest (0.770), followed by YOLO11m and YOLOv10m. YOLOv9c suffered a dramatic drop (0.328).

Under low contrast and blur, the models experienced a drop, with YOLOv12m maintaining the highest values (0.712 and 0.557, respectively). The instability of Yolov9c continued.
The most damaging distortion on unseen data is noise. Even YOLOv12m dropped to 0.192, reinforcing that exposure to noise during training may be essential for robustness. YOLOv10 and 11 performed similarly here (~0.145), showing a universal weakness in handling high-frequency disruptions without prior exposure. YOLOv12m and YOLO11m handled both bluish and greenish shifts relatively well (0.739 and 0.759), showing strong adaptability to common underwater tints.
Again, YOLOv12m showed better performance (0.649) under clean-water conditions, followed by YOLOv10m (0.638).
In summary, YOLOv12m consistently outperformed others in both familiar and unfamiliar settings, noise remained the most harmful distortion, in unseen data. Color distortions were relatively less harmful compared to structural degradations such as blur, clean-water, and noise.

\begin{table*}[htbp]
\centering
\caption{Comparative mAP@50 performance across YOLO models under different underwater distortions on unseen images}
\label{Table.4}
\begin{adjustbox}{width=\textwidth}
\begin{tabular}{lccccccc}
\toprule
\textbf{Model} & \textbf{Original} & \textbf{Low-Contrast} & \textbf{Blur} & \textbf{Noise} & \textbf{Bluish} & \textbf{Greenish} & \textbf{Clean-Water} \\
\midrule
YOLOv8m   & 0.758 & 0.712 & 0.534 & 0.181 & 0.733 & 0.745 & \textbf{0.652} \\
YOLOv9c   & 0.328 & 0.336 & 0.262 & 0.176 & 0.367 & 0.350 & 0.311 \\
YOLOv10m  & 0.746 & 0.682 & 0.517 & 0.145 & 0.706 & 0.737 & 0.638 \\
YOLO11m   & 0.752 & 0.691 & 0.498 & 0.143 & 0.714 & 0.742 & 0.628 \\
YOLOv12m  & \textbf{0.770} & \textbf{0.712} & \textbf{0.557} & \textbf{0.192} & \textbf{0.739} & \textbf{0.759} & 0.649 \\
\bottomrule
\end{tabular}
\end{adjustbox}
\end{table*}

\subsection{Feature-Level Sensitivity (RQ2)}

\paragraph{Experimental Recap}
This study examined how low-level image features in different underwater image conditions influence detection performance. Texture-based features from the Gray-Level Co-occurrence Matrix: \textit{contrast}, \textit{correlation}, \textit{energy}, and \textit{homogeneity} were calculated for three spatial scopes:
\begin{itemize}
    \item \textbf{full}: Entire scene, combining object and background textures.
    \item \textbf{obj}: Annotated object regions only.
    \item \textbf{bg}: Background regions only, capturing contextual textures like seabed structures, water particles, or smooth open water.
\end{itemize}

\begin{figure*}[htbp]
    \centering
    \includegraphics[width=0.7\textwidth]{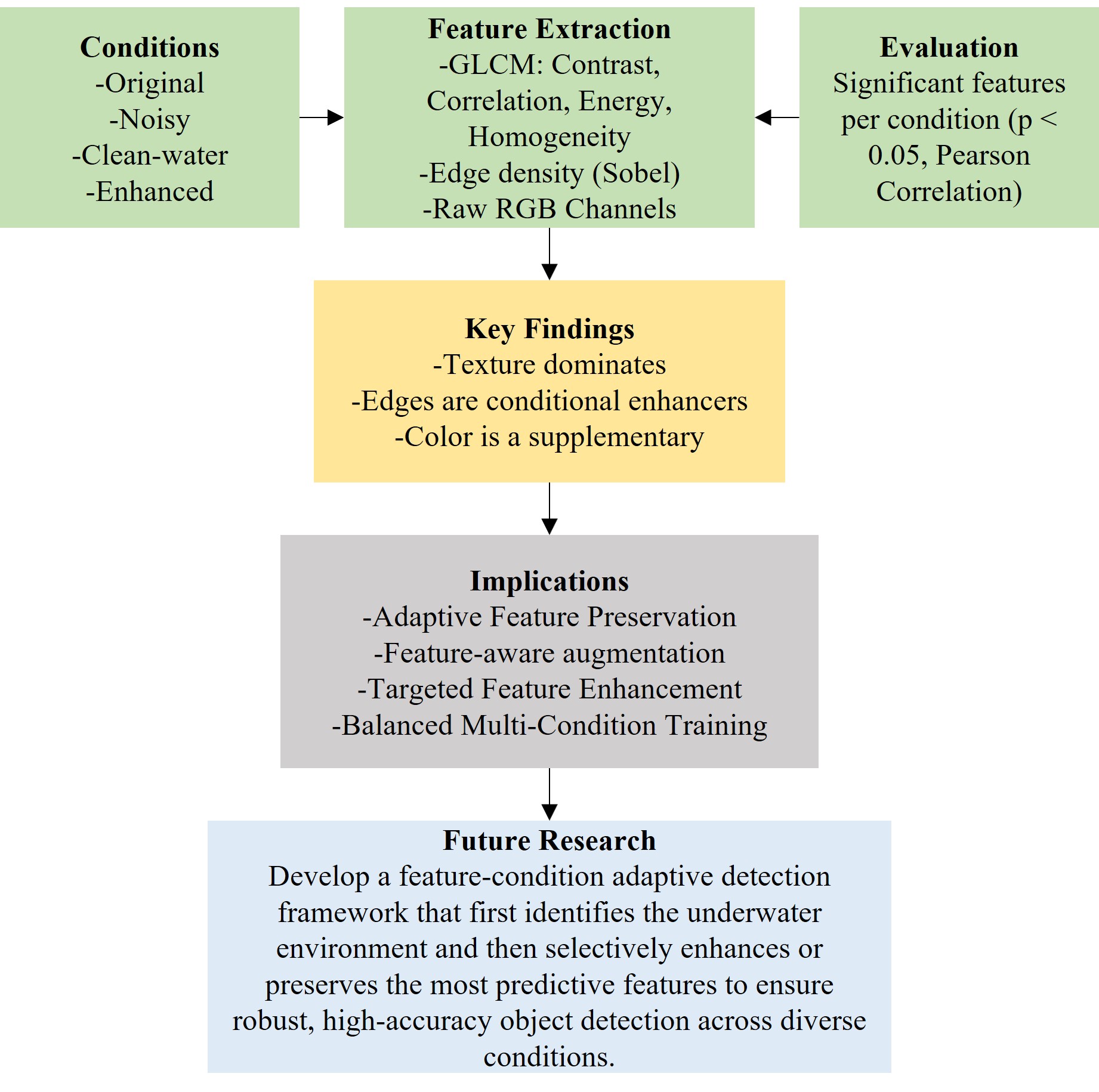} 
    \caption{Underwater Object Detection Low-Level Feature Analysis Flow and Results}
    \label{Fig.5}
\end{figure*}

In addition to texture, edge densities were extracted using Sobel filters, and color values were derived directly from the raw RGB channels to compare their impact alongside texture cues. Figures\ref{Fig.5} shows the flow and Table~\ref{Table.5} presents the mean feature values across the various underwater image conditions examined in this study.

\paragraph{Result Observations and Interpretation}

GLCM features capture complementary aspects of image structure: (i) \textbf{Contrast} measures local intensity variation, where meaningful edges aid localization, but noise-induced contrast can mislead detection; (ii) \textbf{Correlation} reflects the predictability of neighboring pixel values, with consistent patterns within objects supporting classification, but highly correlated backgrounds potentially camouflaging targets; (iii) \textbf{Energy} measures texture uniformity, where smooth object textures simplify recognition, while overly complex backgrounds may cause confusion; and (iv) \textbf{Homogeneity} indicates the similarity of neighboring pixel intensities, with smooth backgrounds enhancing object visibility, but excessive uniformity in objects reducing distinctiveness.

\begin{table*}[htbp]
\centering
\caption{Comparison of Feature Values (mean ± standard deviation) across Different Image Conditions}
\label{Table.5}
\footnotesize 
\setlength{\tabcolsep}{6pt} 
\begin{tabular}{llcccc}
\toprule
\textbf{Feature} & \textbf{Feature Type} & \textbf{Original} & \textbf{Clean-Water} & \textbf{Noise} & \textbf{Enhanced} \\
\midrule
\multirow{3}{*}{Color} 
  & Mean\_R & 21.814 & 27.053 & 21.107 & 24.446 \\
  & Mean\_G & 27.928 & 31.947 & 27.878 & 25.682 \\
  & Mean\_B & 23.399 & 28.904 & 23.197 & 22.873 \\
\midrule
\multirow{12}{*}{Texture}
  & Full\_Contrast     & 16.526 & 64.467 & 28.013 & 28.507 \\
  & Full\_Correlation  & 0.009  & 0.0067 & 0.1733 & 0.0088 \\
  & Full\_Energy       & 0.029  & 0.0228 & 0.0017 & 0.0175 \\
  & Full\_Homogeneity  & 0.154  & 0.1607 & 0.0020 & 0.1653 \\
  & Obj\_Contrast      & 32.890 & 47.635 & 75.723 & 40.598 \\
  & Obj\_Correlation   & 0.008  & 0.0142 & 0.0170 & 0.0104 \\
  & Obj\_Energy        & 0.103  & 0.1032 & 0.1032 & 0.1032 \\
  & Obj\_Homogeneity   & 0.058  & 0.0713 & 0.0976 & 0.0720 \\
  & Bg\_Contrast       & 36.320 & 73.375 & 43.898 & 47.289 \\
  & Bg\_Correlation    & 0.007  & 0.0061 & 0.0843 & 0.0067 \\
  & Bg\_Energy         & 0.089  & 0.0949 & 0.1006 & 0.0948 \\
  & Bg\_Homogeneity    & 0.137  & 0.1466 & 0.0953 & 0.1489 \\
\midrule
Edge & Object\_Feature & 0.022 & 0.089 & 0.031 & 0.053 \\
\bottomrule
\end{tabular}
\end{table*}

\begin{figure*}[htbp]
    \centering
    \includegraphics[width=\textwidth]{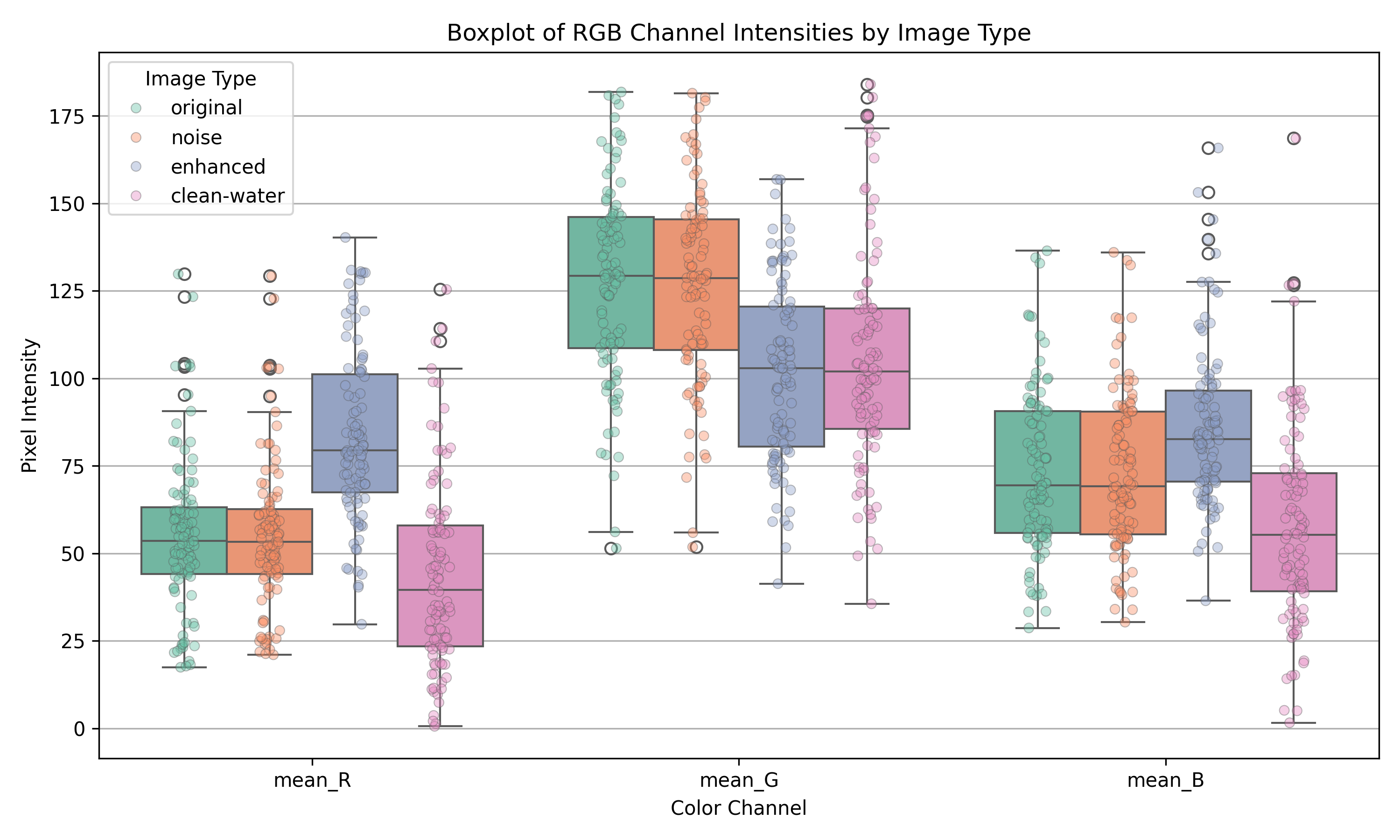}
    \caption{RGB Color Channel Distribution}
    \label{Fig.6}
\end{figure*}

\begin{figure*}[htbp]
    \centering
    \includegraphics[width=\textwidth]{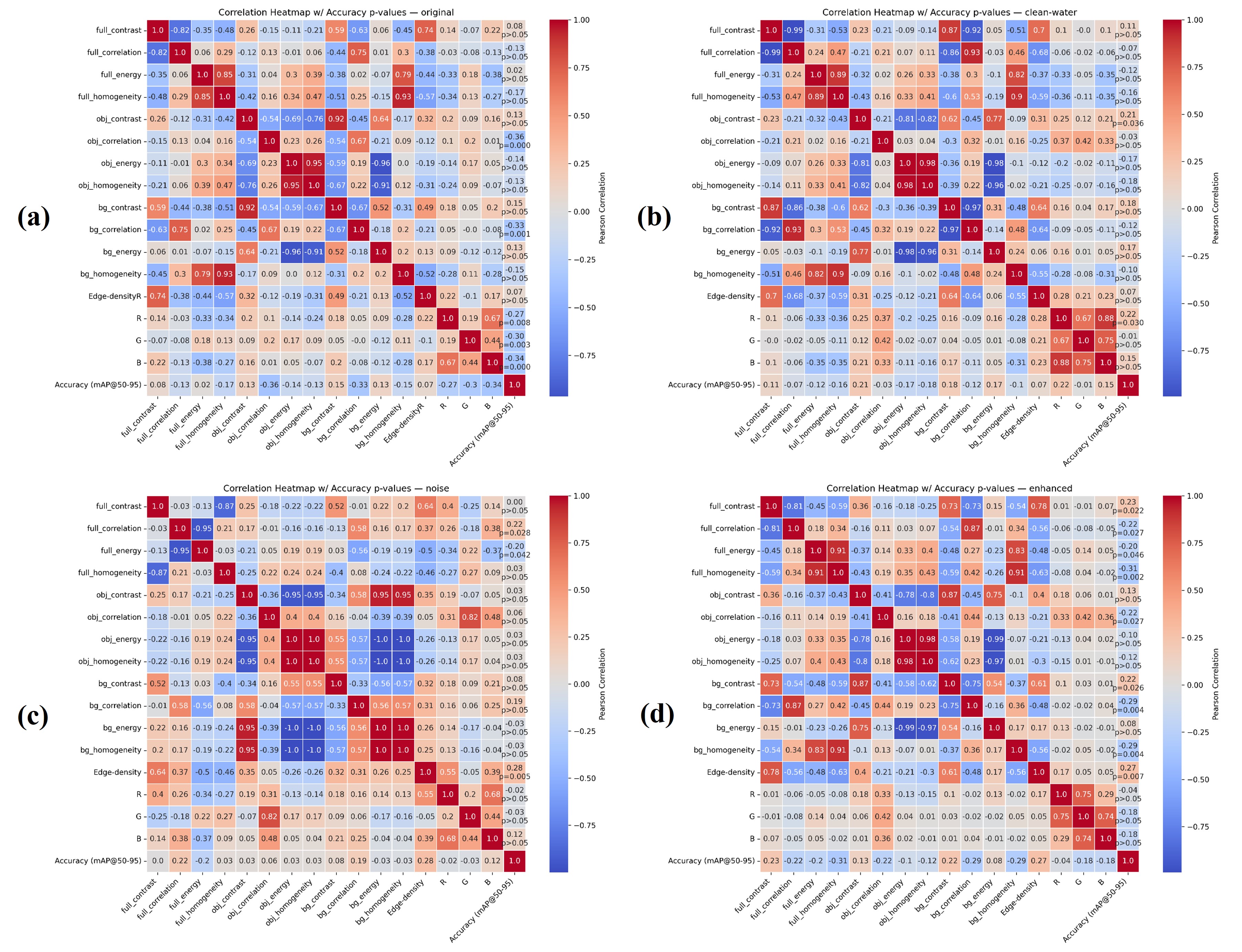}
    \caption{Pearson correlation heatmap showing the relationship between low-level features and detection accuracy (mAP@50–95), with corresponding p-values: (a) original images, (b) clean-water images, (c) noisy images, and (d) enhanced images.}
    \label{Fig.7}
\end{figure*}

The statistical analysis (Pearson correlation, $p < 0.05$) in Figure\ref{Fig.7} revealed that feature importance shifts across imaging conditions, meaning no single feature consistently dominates in all scenarios: (1) \textbf{In original images,} significant features were \textit{object correlation}, \textit{background correlation}, and \textit{R channel}, suggesting that maintaining consistent texture relationships within objects and reducing distracting background patterns supports detection. The R channel’s role may relate to red tones being better preserved in shallower or controlled environments; (2) \textbf{In noisy images,} significant features shifted to \textit{full-correlation}, \textit{full-energy}, and \textit{edge density}, indicating that noise obscures fine object-background boundaries, making global texture relationships and strong edges more important for robustness; (3) \textbf{In clean-water images,} only \textit{object contrast} was significant, showing that in optimal clarity, the primary detection cue is the intensity difference between objects and surroundings; and (4) \textbf{With enhanced images,} a broad set of features became significant (\textit{full-contrast}, \textit{full-energy}, \textit{full-homogeneity}, \textit{full-correlation}, \textit{object correlation}, \textit{background contrast}, \textit{background correlation}, \textit{background homogeneity}, and \textit{edge density}), where enhancement boosts visibility but can alter feature distributions.

\paragraph{Overall Feature Hierarchy}
Across conditions, texture-related features (\textit{contrast}, \textit{correlation}, \textit{energy}, \textit{homogeneity}) emerged as the most consistent predictors of detection accuracy. Edges (\textit{edge density}) played a supportive role under noise and enhancement, while color channels especially red showed limited, condition-specific influence. This hierarchy suggests that texture is primary, edges are conditional enhancers, and color has less influence in underwater detection results.

Figures\ref{Fig.6} shows the distribution of mean pixel intensities for the R, G, and B channels across 100 underwater images under four conditions: original, noisy, enhanced, and clean-water. The boxplots, overlaid with individual points, reveal that the green channel consistently exhibits the highest intensity values across all conditions, reflecting its dominance in underwater imagery due to faster attenuation of red and blue light. The red channel shows strong attenuation in clean-water and original images but increases markedly after enhancement, while the blue channel displays moderate intensities with greater variability in enhanced conditions. These trends highlight condition-dependent color shifts that may influence object detection performance.

\begin{figure*}[htbp]
    \centering
    \includegraphics[width=\textwidth]{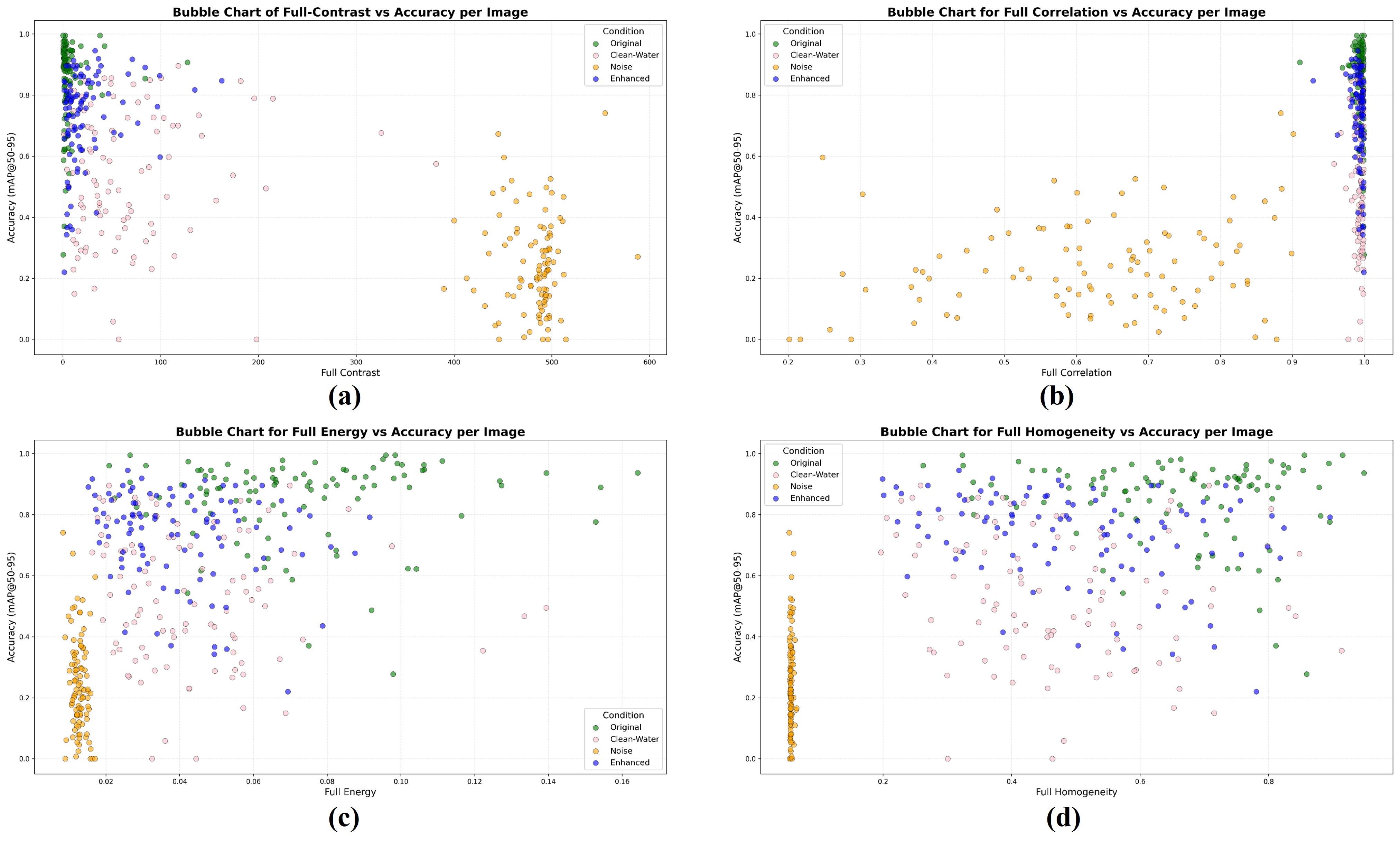}
    \caption{Texture values in terms of Full Contrast, Correlation, Energy and Homogeneity versus Accuracy}
    \label{Fig.8}
\end{figure*}

Figure \ref{Fig.8} illustrates the relationship between four global (full-image) GLCM features: contrast, correlation, energy, and homogeneity and object detection accuracy (mAP@50–95) across 100 underwater images in four conditions: original, clean-water, noisy, and enhanced.
In Figure 8 (a), full-contrast values are generally low for original, enhanced, and clean-water images, with high accuracies clustering in this low range. In contrast, noisy images show very high contrast values but correspondingly low accuracies, suggesting that noise-induced contrast disrupts, rather than supports, detection.
In Figure 8 (b), full-correlation values remain high for most original, enhanced, and clean-water images with high detection accuracy, while noisy images scatter widely with reduced performance. This reinforces the idea that stable texture relationships at a global scale aid detection, but noise breaks this consistency.
Figure 8 (c) shows that full-energy indicating texture uniformity tends to be low for noisy images, which also have poor detection scores. Enhanced and original images show slightly higher energy values associated with better accuracy, while clean-water images cluster at the lowest energy values, possibly due to minimal background texture.
Finally, Figure 8 (d) highlights that higher full-homogeneity values generally align with higher accuracies, particularly in original and enhanced conditions. Noisy images, with low homogeneity, consistently perform poorly, indicating that a smoother global texture distribution benefits detection robustness.

Overall, these plots demonstrate that global texture features (contrast, correlation, energy, and homogeneity) contribute differently depending on the underwater condition. A feature that enhances detection in one scenario for instance, high homogeneity in clean-water images may offer little benefit or even reduce accuracy in another, such as noisy, high-contrast environments. This variability confirms our central argument: low-level visual features degrade unevenly across underwater conditions, and their role in detection performance is highly context-dependent. Consequently, no single feature is universally reliable; understanding these feature-specific vulnerabilities is key to building more robust underwater object detection models.

\begin{figure*}[htbp]
    \centering
    \includegraphics[width=\textwidth]{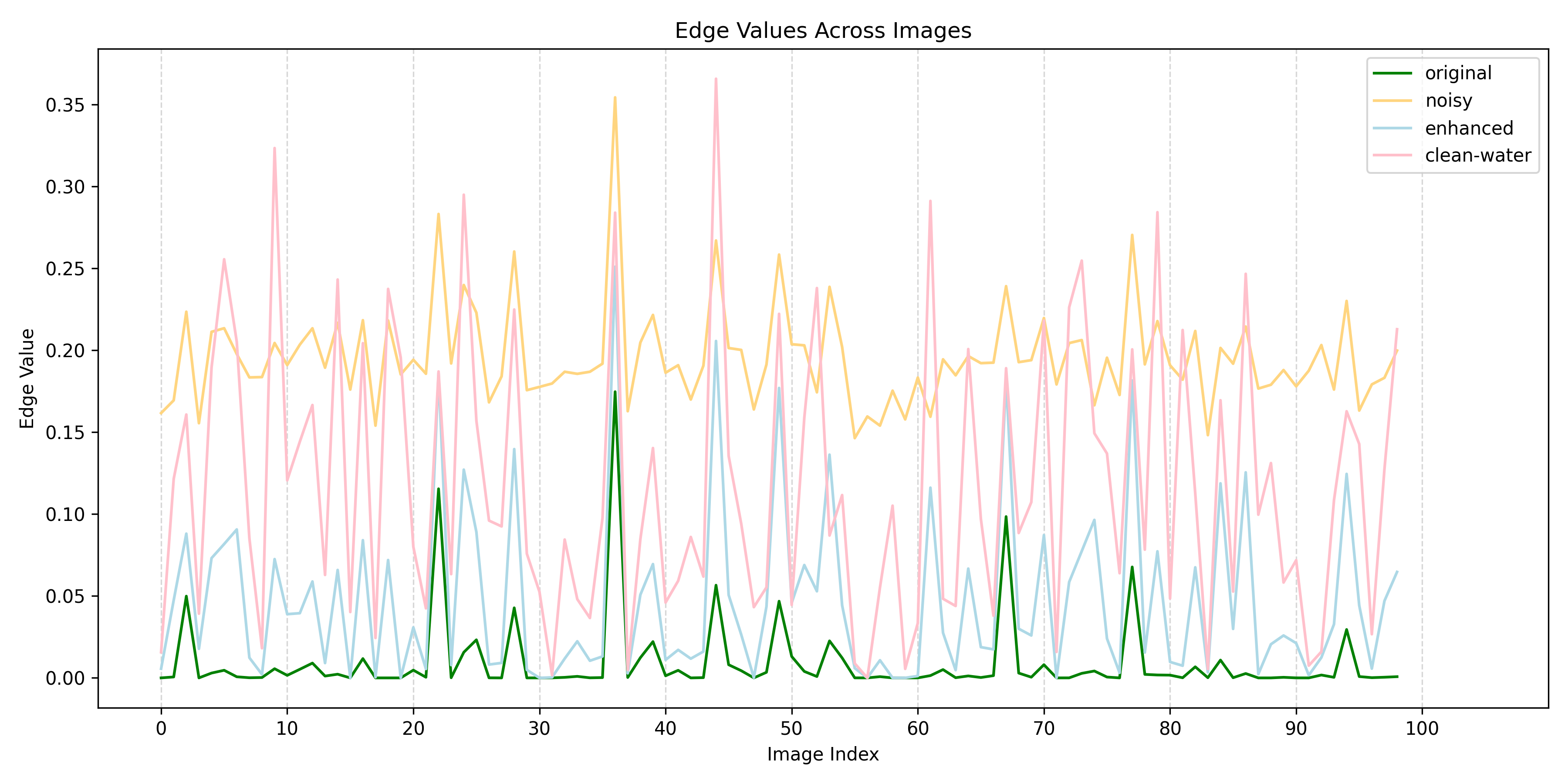}
    \caption{Sobel edge density values across 100 underwater images under four conditions: original (green), noisy (orange), enhanced (blue), and clean-water (pink). Original images show very low edge values, noisy images display artificially high edges, enhanced images moderately increase edge contrast for better detection, while clean-water images show variable but meaningful edges linked to true object boundaries.}
    \label{Fig.9}
\end{figure*}

Figure \ref{Fig.9} presents the variation of edge density values across 100 underwater images for four conditions: original, noisy, enhanced, and clean-water. The original images (green line) consistently exhibit very low edge values, suggesting minimal strong edge structures which is likely due to the inherent blurring and low contrast in raw underwater imagery. On the other hand, noisy images (yellow line) maintain consistently high edge values across almost all images, reflecting the artificial edges introduced by noise patterns rather than meaningful object boundaries.
Enhanced images (blue line) generally have higher edge values than the original set, though still lower than noisy or clean-water conditions, indicating that enhancement boosts edge contrast but not excessively. Clean-water images (pink line) show the highest variability in edge values, with frequent peaks and drops, possibly due to clearer separation between object boundaries and backgrounds under ideal conditions.

Overall, the relationship between edge values and detection accuracy shows that edges alone are not reliable indicators of detection quality. Better detection in the original images, despite their low edge values, suggests that robustness may stem from the model’s training exposure rather than strong edge structures. Enhanced images, with moderate edge values, demonstrate that targeted enhancement can improve detection when carefully designed for feature preservation rather than visual appeal. Clean-water images, which show sharp edges due to simple contrast enhancement, achieve decent but inconsistent results indicating that contrast adjustment must be applied thoughtfully to truly benefit detection. In contrast, noisy images, where high edge values correspond to poor detection, highlight how artificial edges from noise mislead the model and divert it from real object boundaries.

Across all conditions, low-level features; color, texture, and edges, each contribute in different but complementary ways to object detection. Color improvements, particularly the separation of green and blue channels, enhance object-background distinction. Texture features are most effective when they strike a balance between detail and uniformity, helping the model capture fine structures without being overwhelmed by noise. Edge features can be powerful for locating boundaries but must be carefully preserved to avoid confusion from noise-induced artifacts. Ultimately, strong underwater detection relies on models that can jointly leverage these features, ensuring that edges, textures, and colors are preserved and enhanced in ways that support reliable detection under challenging conditions.

\begin{table}[htbp]
\centering
\caption{Low-Level Quality Metrics (↑ indicates higher is better)}
\label{Table.6}
\footnotesize
\begin{adjustbox}{width=\columnwidth}
\begin{tabular}{lccccc}
\toprule
\textbf{Image Type} & \textbf{PSNR (↑)} & \textbf{SSIM (↑)} & \textbf{UIQM (↑)} & \textbf{UCIQE (↑)} & \textbf{URanker (↑)} \\
\midrule
Original  & --     & --    & 0.389 & 0.484 & -0.081 \\
Enhanced  & 18.231 & 0.664 & 0.858 & 0.536 &  1.261 \\
\bottomrule
\end{tabular}
\end{adjustbox}
\end{table}

Table~\ref{Table.6} compares image quality between enhanced images and the original real-world images, used here as ground truth. The enhanced images have lower PSNR (18.23) and SSIM (0.664), indicating notable structural changes from the originals. However, higher UCIQE (0.536) and URanker (1.261) scores suggest improved visual clarity and salience, which can aid object detection in low-visibility underwater scenes if semantic content is preserved. The key trade-off is that excessive processing may remove or distort fine details, potentially reducing model generalization, as noted in \cite{saleem_understanding_2025}. In short, enhancement can boost visibility and detection performance, but only when applied in a way that avoids introducing structural artifacts.

\begin{figure*}[htbp]
    \centering
    \includegraphics[width=\textwidth]{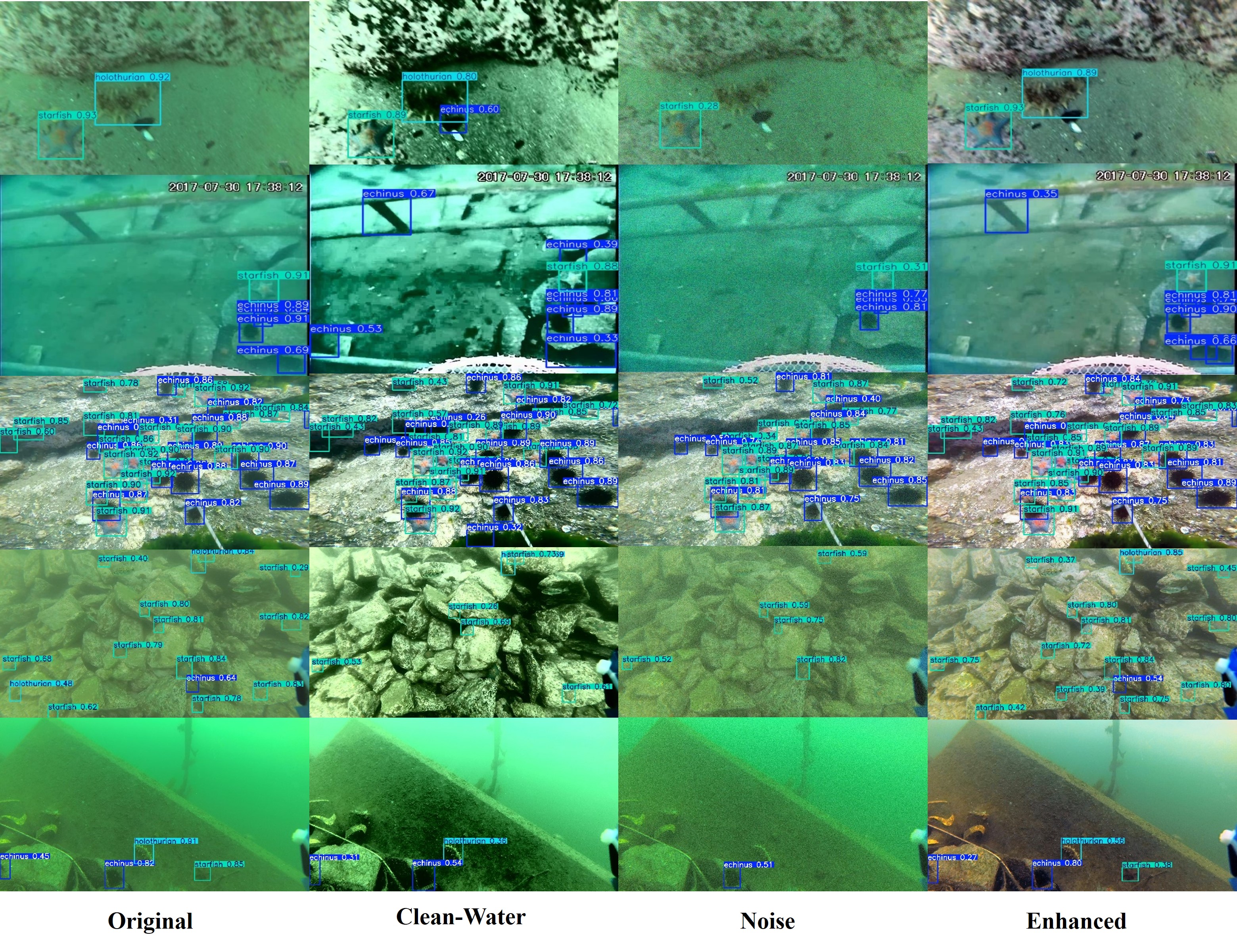}
    \caption{Predictions generated by the YOLOv12 base model on four different conditions of underwater images: (a) Original, (b) Clean-water, (c) Noisy, and (d) Deep-learning enhanced images.}
    \label{Fig.10}
\end{figure*}

Figure \ref{Fig.10} illustrates how detection results change under different underwater image conditions. In the top row, for example, a “holothurian” is detected in the original image with high confidence (0.92) but completely missed in the noise-distorted version. The clean-water version detects it again, though at a lower confidence (0.80), hinting at structural inconsistencies from basic contrast adjustments. In contrast, the enhanced version recovers the detection with a stronger score (0.89), showing how advanced enhancement can restore crucial visual cues for accurate recognition. Similar patterns appear across the rows: \textit{clean-water images can add unintended textures, noisy images hide details, and enhanced images often restore clarity and boost confidence}. These shifts underline how sensitive object detection is to image quality, and how choosing the right enhancement method can make a critical difference in challenging underwater settings.

\paragraph{Research Implications}
\begin{enumerate}
    \item A universal enhancement pipeline is unlikely to work for all environments. Preprocessing should adapt to the most predictive features for each condition.
    \item Data augmentation should deliberately simulate feature variations relevant to certain environments (e.g. edge sharpening for noisy conditions).
    \item Detection models can integrate modules that explicitly reinforce condition-relevant features, such as attention to object contrast in clean water or global correlation in noisy water.
    \item Training on mixed conditions but weighting features according to their condition-specific importance could improve generalization.
\end{enumerate}

Building on these findings, our next research will focus on developing a feature-condition adaptive object detection framework.

\subsection{Impact of Class Frequency and Object Appearance on Detection Performance (RQ3)}

\paragraph{Experimental Recap} 

This experiment set out to understand how class frequency impacts detection performance. Interestingly, even though echinus appeared most frequently (67.8\% of instances), starfish consistently achieved nearly equal mAP scores, as shown in Table~\ref{Table.3}. This pattern prompted a deeper dive: Was it the number of images, the count of object instances, or something about the objects themselves driving this result? \autoref{subsec:Class Imbalance analysis} explains the detailed setup of this experiment.

\paragraph{Result Observations and interpretation}

\begin{table*}[htbp]
\centering
\caption{Performance metrics and data distribution across different training subsets. The training data was divided into seven subsets, each containing a different number of Echinus-containing images: 200, 400, 600, 800, 1000, 1200, and 2000 referred to as Subsets 1 through 7, respectively.}
\label{Table.7}
\footnotesize
\begin{tabular}{llcccccc}
\hline
\textbf{Subset} & \textbf{Class} & \textbf{Img(\%)} & \textbf{Inst(\%)} & \textbf{P} & \textbf{R} & \textbf{mAP@50} & \textbf{mAP@50-95} \\
\hline
\multirow{5}{*}{1}
 & all          & -     & -     & 0.608 & 0.491 & 0.540 & 0.299 \\
 & echinus      & 14.2  & 19.02 & \textbf{0.710} & 0.619 & \textbf{0.683} & 0.372 \\
 & holothurian  & 57.0  & 26.48 & 0.519 & 0.471 & 0.457 & 0.240 \\
 & scallop      & 20.1  & 15.58 & 0.663 & 0.220 & 0.364 & 0.194 \\
 & starfish     & 57.3  & 38.92 & 0.538 & \textbf{0.652} & 0.655 & \textbf{0.389} \\
\hline
\multirow{5}{*}{2}
 & all          & -     & -     & 0.753 & 0.585 & 0.658 & 0.385 \\
 & echinus      & 24.9  & 30.94 & \textbf{0.846} & \textbf{0.708} & \textbf{0.803} & 0.466 \\
 & holothurian  & 56.2  & 22.77 & 0.722 & 0.557 & 0.607 & 0.342 \\
 & scallop      & 18.9  & 12.28 & 0.615 & 0.404 & 0.475 & 0.262 \\
 & starfish     & 58.2  & 34.01 & 0.828 & 0.671 & 0.747 & \textbf{0.468} \\
\hline
\multirow{5}{*}{3}
 & all          & -     & -     & 0.741 & 0.623 & 0.683 & 0.399 \\
 & echinus      & 33.2  & 38.47 & \textbf{0.839} & \textbf{0.766} & \textbf{0.842} & \textbf{0.490} \\
 & holothurian  & 55.2  & 20.69 & 0.610 & 0.631 & 0.626 & 0.357 \\
 & scallop      & 17.2  & 10.21 & 0.725 & 0.390 & 0.503 & 0.273 \\
 & starfish     & 58.0  & 30.62 & 0.790 & 0.705 & 0.763 & 0.478 \\
\hline
\multirow{5}{*}{4}
 & all          & -     & -     & 0.778 & 0.631 & 0.706 & 0.432 \\
 & echinus      & 39.9  & 45.25 & \textbf{0.855} & \textbf{0.780} & \textbf{0.860} & \textbf{0.528} \\
 & holothurian  & 53.0  & 17.47 & 0.766 & 0.576 & 0.646 & 0.392 \\
 & scallop      & 16.3  & 8.44  & 0.690 & 0.431 & 0.524 & 0.298 \\
 & starfish     & 58.0  & 28.84 & 0.801 & 0.735 & 0.796 & 0.510 \\
\hline
\multirow{5}{*}{5}
 & all          & -     & -     & 0.765 & 0.636 & 0.715 & 0.435 \\
 & echinus      & 45.3  & 48.08 & \textbf{0.863} & \textbf{0.748} & \textbf{0.851} & \textbf{0.525} \\
 & holothurian  & 52.5  & 17.09 & 0.705 & 0.637 & 0.672 & 0.396 \\
 & scallop      & 15.2  & 7.84  & 0.692 & 0.422 & 0.534 & 0.303 \\
 & starfish     & 57.3  & 27.01 & 0.802 & 0.737 & 0.804 & 0.518 \\
\hline
\multirow{5}{*}{6}
 & all          & -     & -     & 0.780 & 0.662 & 0.730 & 0.451 \\
 & echinus      & 49.9  & 50.45 & \textbf{0.841} & \textbf{0.808} & \textbf{0.869} & \textbf{0.543} \\
 & holothurian  & 51.3  & 16.24 & 0.742 & 0.659 & 0.696 & 0.428 \\
 & scallop      & 14.3  & 7.04  & 0.733 & 0.431 & 0.547 & 0.306 \\
 & starfish     & 57.7  & 26.27 & 0.806 & 0.752 & 0.809 & 0.528 \\
\hline
\multirow{5}{*}{7}
 & all          & -     & -     & 0.771 & 0.640 & 0.716 & 0.445 \\
 & echinus      & 62.4  & 56.20 & 0.836 & \textbf{0.805} & \textbf{0.867} & \textbf{0.547} \\
 & holothurian  & 49.4  & 14.58 & 0.699 & 0.655 & 0.672 & 0.411 \\
 & scallop      & 12.9  & 5.76  & 0.707 & 0.381 & 0.519 & 0.293 \\
 & starfish     & 59.3  & 23.46 & \textbf{0.841} & 0.720 & 0.805 & 0.529 \\
\hline
\end{tabular}
\end{table*}

\begin{figure*}[htbp]
    \centering
    \includegraphics[width=\textwidth]{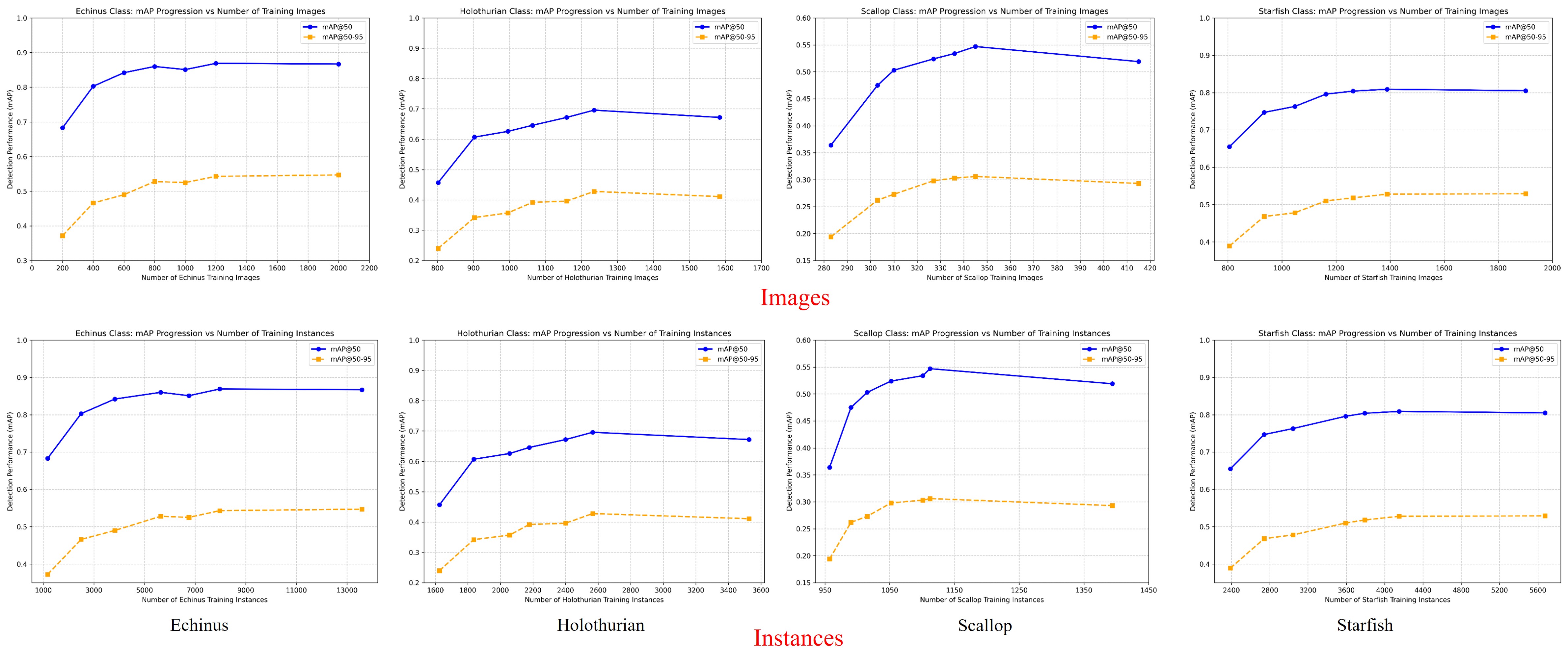}
    \caption{Detection performance progression (mAP@50 and mAP@50–95) for four underwater object classes (echinus, holothurian, scallop, and starfish) as the number of training images (top row) and instances (bottom row) increases. Across all classes, both metrics generally improve with more training data, but performance gains tend to plateau beyond certain thresholds.}
    \label{Fig.11}
\end{figure*}

As seen in Table~\ref{Table.7} and Figure~\ref{Fig.11}, Echinus consistently shows a strong positive link between the number of training images and detection performance. With just 200 images in Subset 1, the mAP@50 starts at 0.683. As more images are added, accuracy steadily climbs reaching 0.869 in Subset 6 and peaking at 0.867 in Subset 7. This trend clearly highlights how having more examples helps the model learn class-specific features more effectively. However, the improvement begins to level off between Subsets 6 and 7, likely due to a slight dip in image count, as reflected in Table~\ref{Table.3}.

Interestingly, Starfish maintains high mAP@50 across all subsets, ranging from 0.655 in Subset 1 to 0.809 in Subset 6 and 0.805 in Subset 7. It also achieves competitive performance with Echinus in mAP@50-95 despite no deliberate increase in its image count. This suggests that Starfish benefits either from clearer visual features such as its distinctive shape and texture or from its consistent co-occurrence with Echinus, which may allow the model to learn shared contextual cues. Overall, these results highlight the importance of object distinctiveness and robustness in appearance, not merely the frequency of occurrence, in driving reliable detection performance.

Holothurian shows a moderate improvement in detection with increasing Echinus images. Its mAP@50 rises from 0.457 in Subset 1 to 0.696 in Subset 6, then slightly drops to 0.672 in Subset 7. It likely benefits indirectly from general improvements in model training as the dataset becomes richer. However, its more ambiguous or variable appearance may limit how much it can benefit compared to Echinus or Starfish.

Scallop remains the most challenging class to detect, with mAP@50 fluctuating between 0.364 (Subset 1) and 0.547 (Subset 6), then dropping slightly in Subset 7. Despite moderate gains, the performance remains low possibly due to several likely reasons: lowest instance count, minimal co-occurrence, and possibly less distinctive features. This shows that low frequency, combined with low visual salience, significantly hinders model learning.

The performance drops observed in the last subset are attributed to random selection, where the number of images decreased in Subset 7 due to sampling randomness, as shown in Table~\ref{Table.3}.

From this analysis, it is clear that class imbalance in underwater object detection cannot be addressed by frequency alone. Object distinctiveness, visual clarity, and co-occurrence context also play major roles.
Building on these findings, in our next research we will design a targeted augmentation strategy that focuses on strengthening the weaker classes (e.g., scallop, holothurian) by generating visibility-enhanced variations. For rare classes, we will integrate generative diffusion models to produce realistic, condition-aware samples that reflect the texture, shape, and environmental cues observed in this study. Additionally, we will apply a class re-weighting scheme to moderately favour under-represented classes during training, ensuring balanced learning without compromising model stability.

\subsection{Small Sample Injection Strategy}

To enhance detection robustness in challenging underwater environments, we devised a training-aware augmentation strategy grounded in findings from RQ1, which identified noise as the most disruptive simulated condition. Inspired by Shen et al. \cite{shen_noise-aware_2020}, our goal is to improve model resilience to both noise and enhanced visibility without sacrificing performance on real-world underwater data. Given its consistently reliable performance in earlier experiments, YOLOv12m was chosen as the base model. We designed two targeted augmentation pipelines:
\begin{itemize}
    \item \textbf{Noise injection} \\
    A dual-noise strategy was applied, combining diffusion noise and salt-and-pepper noise in two configurations:
    \begin{itemize}
        \item \textit{High Noise:} 34\% real-world/original underwater images, 33\% diffusion noise, and 33\% salt-and-pepper noise.
        \item \textit{Low Noise:} 90\% real-world/original underwater images, 5\% diffusion noise, and 5\% salt-and-pepper noise.
    \end{itemize}

    \item \textbf{Clean-Water injection} \\
    To simulate improved visibility conditions (clean-water images), 10\% of the training data was enhanced using CLAHE (Contrast Limited Adaptive Histogram Equalization) and contrast amplification. Each batch consisted of 90\% original and 10\% clean-water images. 
\end{itemize}

\begin{figure*}[htbp]
  \centering
  \includegraphics[width=\textwidth]{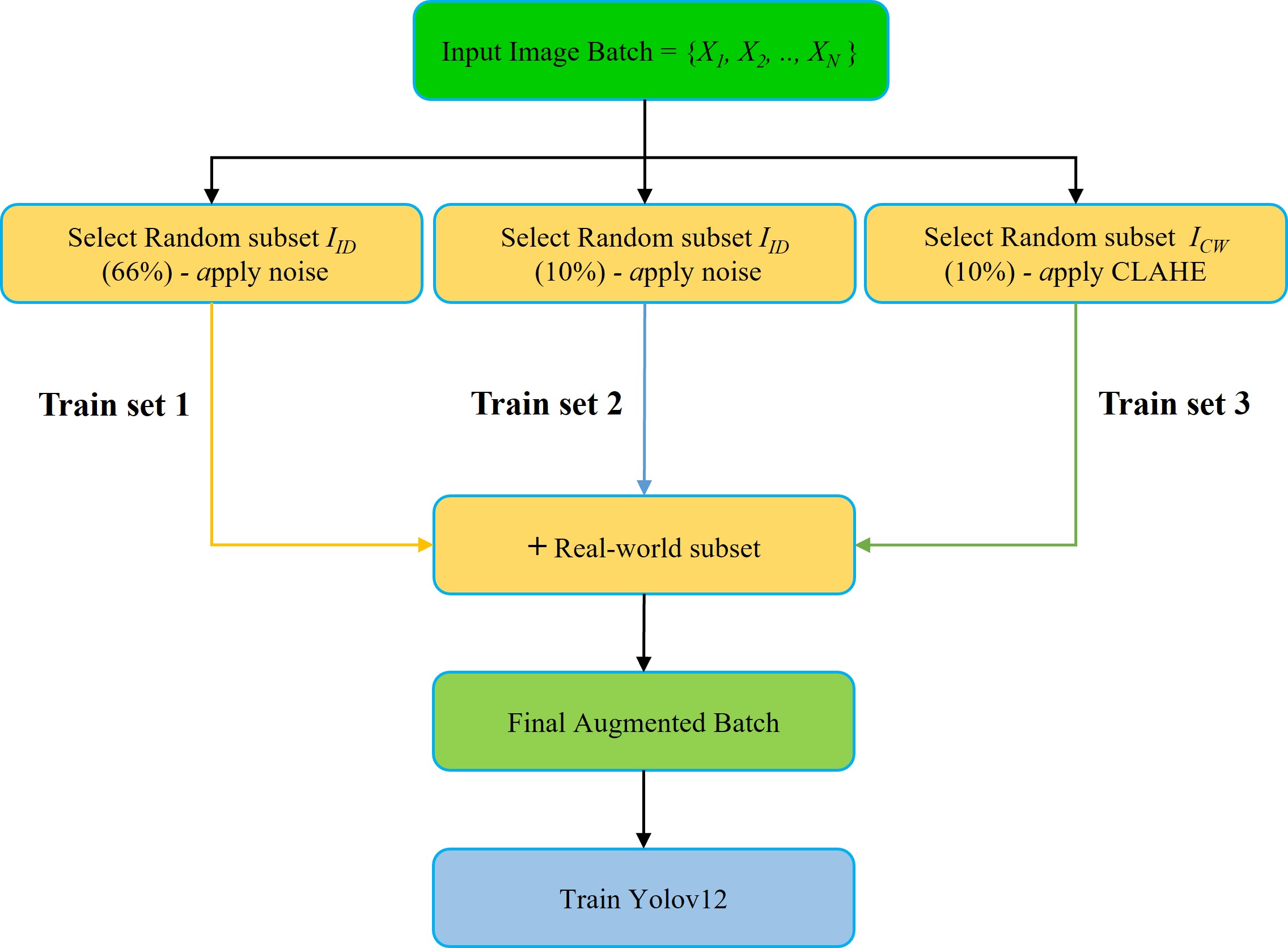}
  \caption{Illustration of the proposed augmentation-aware training strategy for underwater object detection using YOLOv12. Three parallel configurations were explored: (1) High-noise training (66\% noisy + 34\% real-world/original), (2) Low-noise training (10\% noisy + 90\% real-world/original), and (3) Clean-water-aware training (10\% basic-enhanced + 90\% real-world/original).}
  \label{Fig.12}
\end{figure*}

These augmentations were introduced during training as a pre-processing step in the \texttt{train.py} script. All models were trained from scratch using YOLOv12m and evaluated on original, noisy, and clean-water test sets to assess their generalization performance as demonstrated in Figure~\ref{Fig.12}.

We set two targeted augmentation pipelines, resulting in three training types: \textit{Refined Noise}, \textit{High Noise}, and \textit{Clean-Water-Aware} training.
Let a mini-batch be represented as:
\[
\mathcal{B} = \{\mathbf{X}_1, \mathbf{X}_2, \ldots, \mathbf{X}_N\}, \quad \mathbf{X}_i \in \mathbb{R}^{3 \times H \times W}
\]
From each mini-batch, three disjoint random subsets were selected:
\begin{itemize}
    \item $\mathcal{I}_D \subset \mathcal{B}$: images to receive \textbf{diffusion noise}
    \item $\mathcal{I}_{SP} \subset \mathcal{B}$: images to receive \textbf{salt-and-pepper noise}
    \item $\mathcal{I}_{CW} \subset \mathcal{B}$: images to receive \textbf{clean-water enhancement} using CLAHE and contrast amplification
\end{itemize}
The remaining samples were left unaltered:
\[
\mathcal{I}_{clean} = \mathcal{B} \setminus (\mathcal{I}_D \cup \mathcal{I}_{SP} \cup \mathcal{I}_{CW})
\]
Three training configurations were explored:

\vspace{1em}

\textbf{1. High Noise (Strong Perturbation)}  
\begin{equation}
\begin{split}
|\mathcal{I}_D| &= 0.33N, \quad |\mathcal{I}_{SP}| = 0.33N, \\
|\mathcal{I}_{CW}| &= 0, \quad |\mathcal{I}_{real-world}| = 0.34N
\end{split}
\label{eq:high_noise}
\end{equation}

\textbf{2. Low Noise (Mild Perturbation)}  
\begin{equation}
\begin{split}
|\mathcal{I}_D| &= 0.05N, \quad |\mathcal{I}_{SP}| = 0.05N, \\
|\mathcal{I}_{CW}| &= 0, \quad |\mathcal{I}_{real-world}| = 0.90N
\end{split}
\label{eq:low_noise}
\end{equation}

\textbf{3. Clean-Water-Aware Training (Visibility Boost)}  
\begin{equation}
\begin{split}
|\mathcal{I}_D| &= 0, \quad |\mathcal{I}_{SP}| = 0, \\
|\mathcal{I}_{CW}| &= 0.10N, \quad |\mathcal{I}_{real-world}| = 0.90N
\end{split}
\label{eq:clean_water}
\end{equation}

\vspace{1em}
For each $\mathbf{X}_i \in \mathcal{I}_D$, iterative diffusion noise was applied as:
\begin{equation}
\mathbf{X}'_i = \text{clip}\left(\mathbf{X}_i + \sum_{t=1}^{T} \epsilon_t,\ 0,\ 1\right), \quad \epsilon_t \sim \mathcal{N}(0,\ \sigma^2)
\end{equation}
where $\sigma = 0.05$ and $T = 3$ noise steps.

For $\mathbf{X}_j \in \mathcal{I}_{SP}$, 2\% of the pixels were randomly altered with equal probability to either 0 (pepper) or 1 (salt):
\begin{equation}
\mathbf{X}'_j(p) =
\begin{cases}
1, & \text{(salt)} \\
0, & \text{(pepper)}
\end{cases}
\quad \text{with probability } p = 0.01
\end{equation}

For $\mathbf{X}_k \in \mathcal{I}_{CW}$, enhancement was applied using:
\begin{equation}
\mathbf{X}'_k = \text{CLAHE}(\mathbf{X}_k) + \alpha \cdot \text{Contrast}(\mathbf{X}_k), \quad \alpha = 0.5
\end{equation}.

\paragraph{Result Observations and interpretation:}
The presented results in  Table~\ref{Table.8} underscore the significance of training-aware strategies, particularly when addressing underwater image variations such as noise and visibility enhancement. Training-aware methods involve intentionally including a representative proportion of target-domain variations (e.g., noise or enhanced images) in the training data, allowing the model to better generalize during inference.

\paragraph{Noise Injection}
When evaluating the impact of noise-aware training, the following key observations were made:
\begin{itemize}
    \item Training solely on real-world/original data led to a sharp decline under noisy test conditions, with mAP@50:95 falling to 0.151.
    
    \item Introducing large amounts (66\%) (66\%) of noise improved robustness in noisy settings (mAP@50:95 increased to 0.259). However, this came at a cost, as performance on clean images dropped from 0.504 to 0.415, suggesting possible overfitting to distortion patterns.
    
    \item A low noise setup with only 10\% noise yielded the best trade-off, maintaining strong accuracy on real-world/original images (0.505 mAP@50:95) while also boosting robustness in noisy conditions (0.298). This informs us that a modest proportion of noise can effectively improve resilience without compromising generalization.
\end{itemize}

\begin{table*}[htbp]
\centering
\caption{Performance under different training and testing strategies across original/real-world, noise and enhancement settings.}
\label{Table.8}
\footnotesize
\begin{tabularx}{\textwidth}{llXXXX}
\toprule
\textbf{Training Setting} & \textbf{Testing Scenario} & \textbf{Precision (P)} & \textbf{Recall (R)} & \textbf{mAP@50} & \textbf{mAP@50:95} \\
\midrule
\multicolumn{6}{l}{\textbf{Noise}} \\
Trained on Original data   & Original     & 0.808 & 0.690 & 0.770 & 0.504 \\
                           & Noise        & 0.597 & 0.241 & 0.292 & 0.151 \\
Trained on 66\% High noise & Original     & 0.761 & 0.624 & 0.697 & 0.415 \\
                           & Noise        & 0.613 & 0.413 & 0.451 & 0.259 \\
Trained on 10\% Low noise  & Original     & \textbf{0.821} & \textbf{0.685} & \textbf{0.770} & \textbf{0.505} \\
                           & Noise        & \textbf{0.679} & \textbf{0.447} & \textbf{0.512} & \textbf{0.298} \\
\midrule
\multicolumn{6}{l}{\textbf{Clean-Water}} \\
Trained on Original data   & Original     & 0.808 & 0.690 & 0.770 & 0.504 \\
                           & Clean-Water  & 0.789 & 0.567 & 0.649 & 0.408 \\
                           & Enhanced     & \textbf{0.903} & \textbf{0.759} & \textbf{0.842} & \textbf{0.631} \\
Trained on 10\% clean-water& Original     & 0.791 & 0.693 & 0.756 & 0.495 \\
                           & Clean-Water  & 0.764 & 0.603 & 0.669 & 0.421 \\
\midrule
\multicolumn{6}{l}{\textbf{Domain Adaptation}} \\
Fine-tuned on Enhanced Dataset & Original & 0.772 & 0.595 & 0.672 & 0.395 \\
                               & Enhanced & \textbf{0.902} & \textbf{0.772} & \textbf{0.899} & \textbf{0.717} \\
\bottomrule
\end{tabularx}
\end{table*}

\paragraph{Clean-water sample injection}
The effect of clean-water sample injection was less consistent compared to noise sample injection:
\begin{itemize}
    \item Training on original data generalized moderately well to CLAHE-enhanced images, achieving 0.408 mAP@50:95.
    
    \item Introducing 10\% clean-water data into the training set yielded only marginal improvements (0.421 mAP@50:95), indicating limited gains compared to baseline.
    
    \item When trained on original data and tested on Hybesense-enhanced images, performance reached 0.842 mAP@50 and 0.631 mAP@50:95, demonstrating stronger generalization to advanced enhancement techniques.
\end{itemize}

\paragraph{Fine-tuning for Domain Shift Adaptation}

Underwater environments are highly diverse, and models trained in one domain (e.g., natural oceans) may fail to generalize when deployed in another (e.g., lakes, dams, or turbid coastal waters). This inevitable domain shift poses a serious barrier for robust underwater object detection. To address this, we explored fine-tuning as a lightweight yet effective adaptation strategy, using advanced enhancement to simulate realistic target conditions.
\begin{itemize}
\item Instead of retraining from scratch, we fine-tuned a pre-trained YOLOv12 model with only 10\% of enhanced training data generated using the HybSense method proposed by Guo et al. \cite{guo_underwater_2025}.
\item Fine-tuning yielded the highest accuracy on enhanced test images, reaching 0.899 mAP@50, while still maintaining competitive accuracy on original real-world images (0.672 mAP@50) as shown in Table~\ref{Table.8}.
\item Despite the small data fraction, the model adapted effectively, showing that \textbf{exposure to a targeted subset of the new domain is sufficient for meaningful transfer}.
\item Compared to full retraining, fine-tuning required far fewer resources and training time, making it a cost-efficient solution for adapting detectors to new underwater conditions.
\end{itemize}
These results demonstrate that fine-tuning is not merely a performance booster, but a \textbf{practical pathway for domain adaptation in underwater vision}. In scenarios where models are trained in one environment but deployed in another, fine-tuning offers an efficient means of transferring knowledge with limited new data, ensuring reliable detection across diverse underwater domains.

\textit{For underwater object detection, mAP@50 is often adopted as a more forgiving evaluation metric, since the complex and distortion-prone underwater environment makes stricter thresholds less representative of practical detection performance}.

\section{Unpacking Robustness and Conclusion}
\subsection{Summary of Findings}
This study systematically evaluated the robustness and adaptability of recent YOLO models when exposed to diverse underwater distortions. We hypothesize that convolutional object detectors are inherently sensitive to high-frequency distortions like noise due to their reliance on spatial coherence and structured feature extraction \cite{ziyadinov_noise_2022}. The findings from RQ1 revealed that although all YOLO variants performed well under original, clean conditions, their performance varied significantly under distortion. YOLOv12 consistently outperformed the others across most distortion types, especially under low contrast, blur, and color shifts. However, noise emerged as the most challenging distortion, with all models including YOLOv12 experiencing substantial drops in accuracy unless specifically trained on noisy data.

In RQ2, we conducted a detailed analysis of how underwater visual distortions specifically noise, clean-water contrast enhancements, and advanced deep learning-based enhancements affect foundational image features and influence object detection performance. The results reveal a clear vulnerability of convolutional neural networks (CNNs) to high-frequency noise, which disrupts their dependency on low-frequency spatial patterns that are crucial for robust feature extraction. This disruption leads to degraded internal representations and a notable drop in detection accuracy, consistent with previous findings on the noise sensitivity of CNN architectures \cite{rodriguez-rodriguez_impact_2024, ziyadinov_noise_2022}. Among the three examined features color, texture, and edges, noise most severely impacted edges and texture, introducing artificial contours, breaking global texture uniformity, and increasing intra-class variability in ways that confuse detection models.
Conversely, our analysis shows that enhancement, particularly deep learning-based approaches jointly optimized with object detection objectives, can preserve or even amplify key visual cues. This enhancement not only improves edge sharpness and object boundary definition but also restores texture patterns that contribute to better object separability in feature space. This aligns with the conclusions of \cite{fu_joint_2023}, where multi-task learning architectures such as DPNet demonstrated superior performance by jointly learning enhancement and detection in a cooperative framework. However, we also observed that conventional “clean-water” contrast adjustments, while visually pleasing, often fail to recover semantic features necessary for reliable detection sometimes introducing inconsistencies that reduce confidence scores.

In RQ3, we explored the role of training data distribution and visual object characteristics in detection performance. Our results support the hypothesis that detectability is jointly determined by instance frequency, co-occurrence, and visual distinctiveness. Echinus demonstrated the expected performance gains from increased sample size, while Starfish maintained high detection accuracy despite fewer instances, likely due to its distinct shape, texture, and strong contrast. Conversely, Scallop, a small, low-contrast class showed consistently lower detection rates, even when present in the dataset. These findings echo \cite{gu_systematic_2023}, which highlighted that object size and visual prominence often outweigh raw instance count in determining detection success.
From this, it is clear that class imbalance in UOD cannot be solved through frequency balancing alone. Object distinctiveness, clarity, and contextual co-occurrence must also be addressed.

\subsection{Limitations and Future Research}

Although this study has yielded valuable insights into the robustness of underwater object detection, several limitations remain:
\begin{itemize}
    \item Although multiple distortion types were considered, these were synthetically generated and may not fully capture the complex, multi-layered degradations encountered in real underwater environments.
    \item The experiments were conducted on a fixed set of YOLO architectures, meaning that conclusions about model robustness may not directly extend to other detection paradigms such as transformer-based or diffusion-based architectures.
    \item The dataset employed contained only four object classes, potentially limiting the generalizability of our findings to the broader diversity of underwater species.
\end{itemize}
Building on these findings, our next phase of research will address both visual-level robustness and data-level imbalance in a unified framework:

\begin{itemize}
    \item \textbf{Visual-level robustness:} We will develop a \textit{feature-condition adaptive detection pipeline} capable of estimating environmental conditions (e.g., noise, turbidity, enhancement artifacts) and selectively enhancing or preserving the most predictive features for that scenario.
    \item \textbf{Data-level balance and augmentation:} We will implement a targeted \textit{class-aware augmentation strategy} informed by the class-specific trends identified in this study: (1) Using generative diffusion models to create realistic, condition-aware samples for rare or visually ambiguous classes, replicating authentic underwater textures and environmental cues, (ii) Applying class re-weighting schemes such as weights or class-balanced focal loss to ensure underrepresented classes receive sufficient learning emphasis without destabilizing the training process.
    \item \textbf{Model expansion:} In parallel, we will explore integrating \textit{Large Vision-Language Models (LVLMs)} and advanced detection architectures, leveraging their ability to combine visual and contextual reasoning for improved adaptability in complex underwater environments.
\end{itemize}
Collectively, these advancements aim to move beyond dataset expansion toward intelligent, condition and class-aware robustness strategies, positioning future models to perform reliably in real-world, heterogeneous underwater settings.

\subsection{Conclusion}
This study investigated the robustness of YOLO-based detectors in underwater environments, focusing on why performance drops and how to address it. Our findings show that robustness is shaped by two key factors: \textit{visual-level feature degradation} and \textit{data-level class imbalance}. Noise proved the most disruptive, severely degrading texture and edge cues, while basic enhancement improved visual appeal but failed to restore semantic features. Deep learning-based enhancement, especially when jointly optimized with detection, preserved critical cues and improved robustness.
Class-wise analysis revealed that detection depends not only on sample quantity but also on object-specific visual properties. Injecting a small amount of noise into training improves the model’s ability to generalize across both real-world and noisy conditions. Similarly, enhancement (especially the recent advanced approaches), when designed to emphasize features relevant for detection, leads to stronger results in underwater settings. Importantly, fine-tuning with enhanced images demonstrates clear value for \textbf{domain shift adaptation}: rather than retraining from scratch, lightweight fine-tuning enabled the model to transfer knowledge efficiently to a new domain, achieving high accuracy with limited additional data while maintaining competitiveness on the original domain. This is especially relevant for real deployments where underwater environments naturally differ (e.g., lakes vs. oceans, shallow vs. deep waters, natural vs. man-made settings).
To address these issues, we advocate moving toward an integrated robustness strategy that combines condition-aware feature preservation with class-aware augmentation and re-weighting. At the visual level, a feature-condition adaptive detection pipeline can selectively enhance or preserve features most predictive for a given scenario, guided by pre-analysis modules that extract texture descriptors. At the data level, targeted augmentation using visibility adjustments, background diversification, and generative diffusion models can enrich rare or visually ambiguous classes, supported by re-weighting schemes to ensure balanced learning.
In sum, our findings suggest that robust underwater object detection cannot be achieved through data expansion or enhancement alone. Instead, it requires an adaptive, dual-level approach that accounts for environmental variability and class imbalance while leveraging fine-tuning as a practical tool for domain adaptation. By integrating these insights into future model designs, detection systems can move closer to maintaining high accuracy and generalization across the heterogeneous, unpredictable conditions of real-world underwater environments.



\subsection*{Statements and Declarations}

During the preparation of this work the author(s) used ChatGPT in order to correct the grammar. After using this tool, the author(s) reviewed and edited the content as needed and take(s) full responsibility for the content of the published article.

\subsection*{Statements and Declarations}
Not Applicable

\subsection*{Consent for publication}
Not Applicable

\subsection*{Declaration of conflicting interest}
The authors declared no potential conflicts of interest with respect to the research, authorship, and/or publication of this article

\subsection*{Funding Statement}
This work was supported by National Natural Science Foundation of China (No. 61972240).

\subsection*{Data Availability}
The data that support the findings of this study will be available upon request.

\bibliographystyle{SageV}     
\bibliography{Main-Manuscript}    


\end{document}